\documentclass[10pt,twocolumn,letterpaper]{article}

\usepackage{cvpr}              % To produce the CAMERA-READY version
\usepackage{booktabs}
\usepackage[table,xcdraw,dvipsnames]{xcolor}
\usepackage{multirow}
\usepackage{lipsum}
\usepackage{adjustbox}
\usepackage{colortbl}
\usepackage{amssymb}
\usepackage{pifont}
\usepackage[accsupp]{axessibility} % Improve readability

\newcommand{\prannay}[1]{{\textcolor{red}{Prannay: #1} }}
\newcommand{\zhizhong}[1]{{\textcolor{blue}{Zhizhong: #1} }}
\newcommand{\hao}[1]{{\textcolor{magenta}{Hao: #1} }}
\newcommand{\yoni}[1]{{\textcolor{olive}{Yoni: #1} }}

\renewcommand{\prannay}[1]{}
\renewcommand{\zhizhong}[1]{}
\renewcommand{\hao}[1]{}
\renewcommand{\yoni}[1]{}

\newcommand{\throne}{\textsc{THRONE}\xspace}

\newcommand{\xmark}{\ding{55}}%
\newcommand{\texttts}[1]{\texttt{\footnotesize #1}}

\definecolor{cvprblue}{rgb}{0.21,0.49,0.74}
\usepackage[pagebackref,breaklinks,colorlinks,citecolor=cvprblue]{hyperref}

\def\paperID{12964} % *** Enter the Paper ID here
\def\confName{CVPR}
\def\confYear{2024}

\title{
    \throne: An Object-based Hallucination Benchmark for \\ the Free-form Generations of Large Vision-Language Models
    \vspace{-0.5cm}
    }

\author{Prannay Kaul$^{1}\thanks{Work conducted during an internship at Amazon}$ \ \ Zhizhong Li$^{2}\thanks{Corresponding author}$ \ \ Hao Yang $^{2}$ \ \ Yonatan Dukler $^{2}$ \\ Ashwin Swaminathan $^{2}$ \ \ C. J. Taylor $^{2}$ \ \ Stefano Soatto $^{2}$ \\
VGG, University of Oxford$^{1}$ \ \ AWS AI Labs$^{2}$ \\
{\tt\small \href{mailto:prannay@robots.ox.ac.uk}{prannay@robots.ox.ac.uk}} \ \ \tt\small \{\href{mailto:lzhizhon@amazon.com}{lzhizhon},\href{mailto:haoyng@amazon.com}{haoyng},\href{mailto:dukler@amazon.com}{dukler},\href{mailto:swashwin@amazon.com}{swashwin},\href{mailto:taylorcj@amazon.com}{taylorcj},\href{mailto:soattos@amazon.com}{soattos}\}@amazon.com
}

\begin{document}
\maketitle
\begin{abstract}
% Despite the impressive progress of large vision-language models (LVLMs),
% their generated text is not always visually grounded,
% demonstrated by hallucinations similar to the large language models (LLMs) they are built upon.
Mitigating hallucinations in large vision-language models (LVLMs) remains an open problem.
Recent benchmarks do not address hallucinations in open-ended free-form responses, which we term ``Type I hallucinations''. 
Instead, they focus on hallucinations responding to very specific question formats---typically a multiple-choice response regarding a particular object or attribute---which
we term ``Type II hallucinations''. Additionally, such benchmarks often require external API calls to models which are subject to change.
In practice, we observe that a reduction in Type II hallucinations does not lead to a reduction in Type I hallucinations but rather that the two forms of hallucinations are often anti-correlated.
To address this, we propose \throne,
a novel object-based automatic framework for quantitatively evaluating Type I hallucinations in LVLM free-form outputs.
We use public language models (LMs) to identify hallucinations in LVLM responses and compute informative metrics.
By evaluating a large selection of recent LVLMs using public datasets, we show that an improvement in existing metrics do not lead to a reduction in
Type I hallucinations, 
and that established benchmarks for measuring Type I hallucinations are incomplete.
Finally, we provide a simple and effective data augmentation method to reduce Type I and Type II hallucinations as a strong baseline. Code is now available at \href{https://github.com/amazon-science/THRONE}{https://github.com/amazon-science/THRONE}.
% We show all evaluated models demonstrate Type I hallucations, \zhizhong{$\leftarrow$ this is well-known, right?}
% \yoni{that} advances in existing metrics for performance are orthogonal\prannay{do not generalize to, change everywhere} to \throne,
% \zhizhong{Are we able to make a more general claim that all popular benchmarks are ill-suited?}

% \zhizhong{Let's move these to the end of intro. I also think 1 and 2 are one, and 4 can be extended to include more findings, like how type i is orthogonal to type ii (also merge second half of 3 there).}
% \prannay{Once we have decided on a name, we can stop saying ``our benchmark'' to reduce word count.}
% \yoni{Does it potentially makes sense to mention Type I/ Type II in the abstract?}
\vspace{-0.2cm}
\end{abstract}
% Established benchmarks in the vision-language field like COCO captions and VQA use handcrafted evaluation rules and metrics
% and are not suited to the improved text style of current LVLMs.
% Recent ``comprehensive'' benchmarks for LVLMs such as MME and MMBench require simple model answers---yes/no or multiple choice lists, respectively,
% often rely on opaque paid API services such as GPT-4 for evaluation, reducing access and transparency, and do not focus on hallucinations. 
% To address these shortcomings, we propose [insert name here], a novel benchmark constructed from a general framework for evaluating hallucinations
% in long text generations from LVLMs.
% We make three contributions:
\cvprsection{Introduction}\label{sec:intro}

\begin{figure*}[t]
  \centering
  \begin{minipage}[!ht]{0.49\linewidth}
      \includegraphics[width=\linewidth]{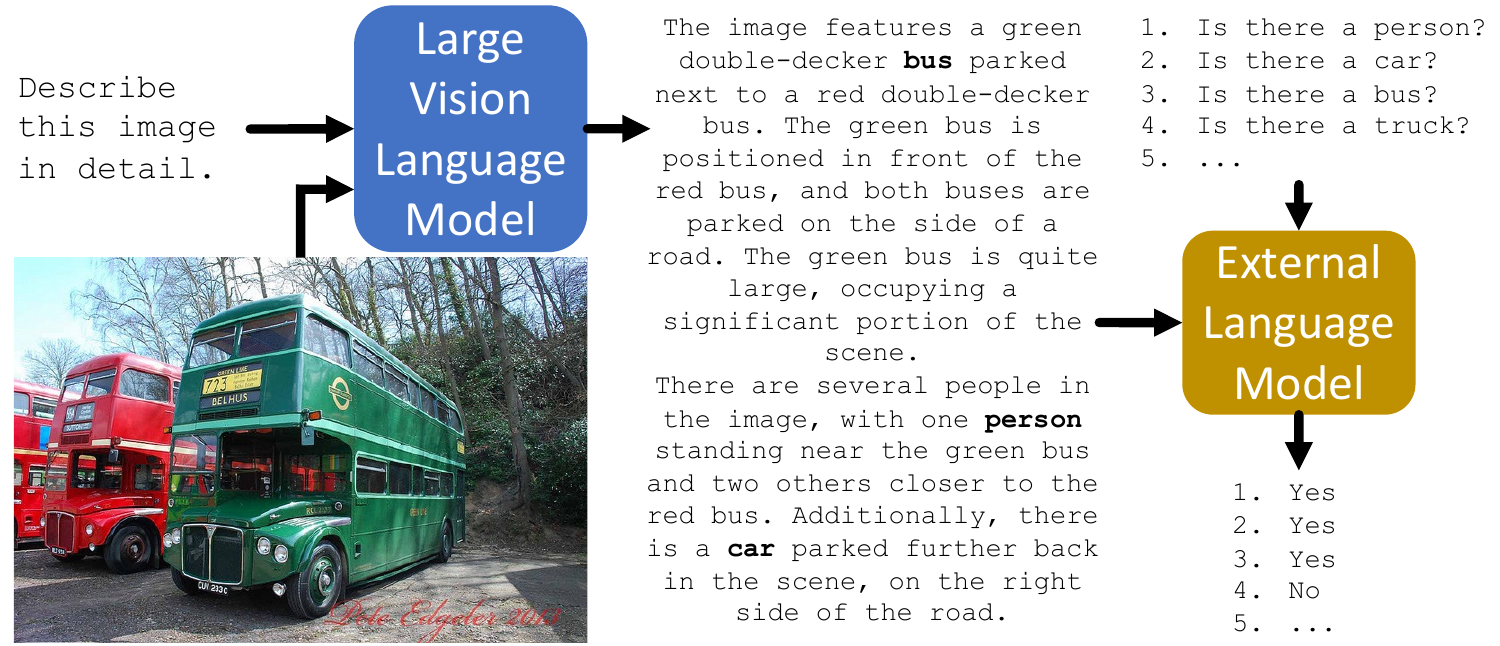}
      \captionof{figure}{\textbf{\throne (Ours):} LVLMs are prompted with a concept neutral instruction.
      An external LM performs abstractive QA on the response to establish the existence of \textbf{Type I} hallucinations.
      }\label{fig:framework_ours}
  \end{minipage}
  \hfill
  \begin{minipage}[!ht]{0.49\linewidth}
      \includegraphics[width=\linewidth]{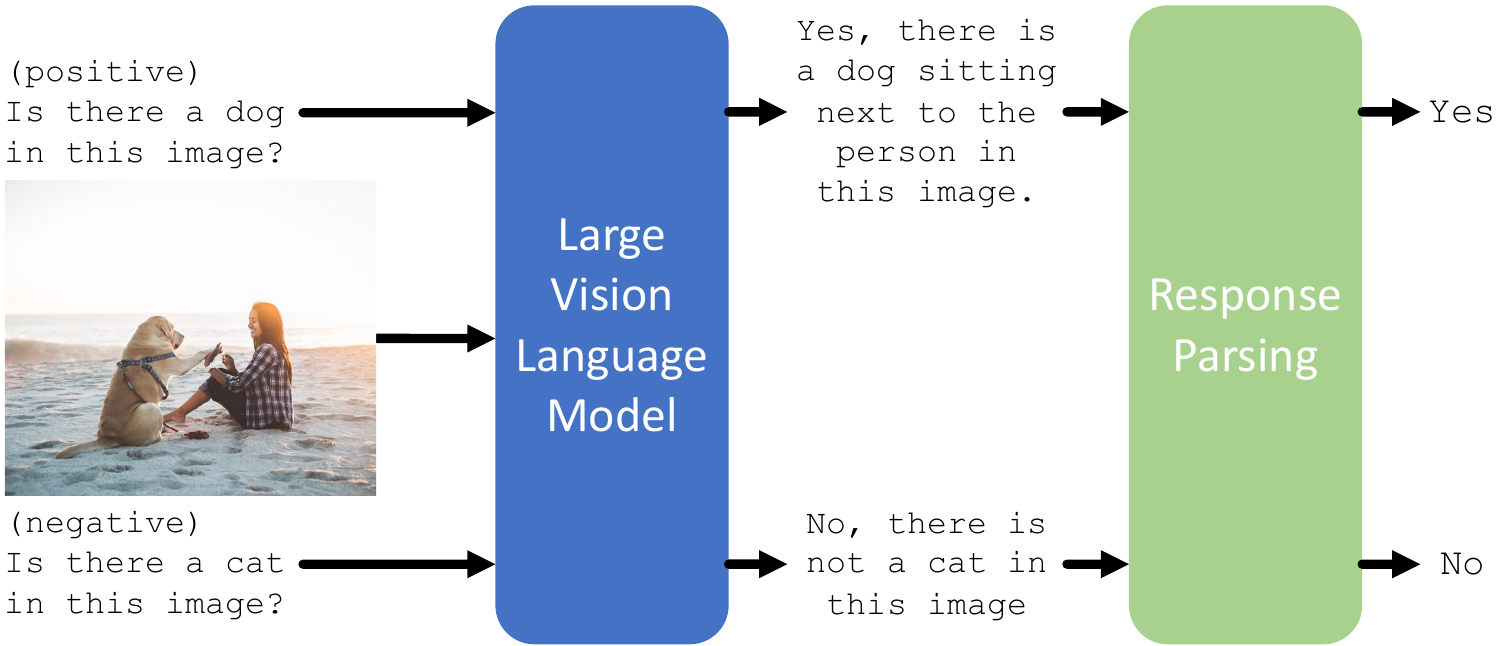}
      \captionof{figure}{\textbf{POPE:} Questions with specific concepts prompt an LVLM directly to evaluate \textbf{Type II} hallucinations~\cite{li2023evaluating}.
      Hand-crafted rules parse LVLM responses to give yes/no labels.
      }\label{fig:framework_pope}
  \end{minipage}

\vspace{-0.5cm}
\end{figure*}

% In this paper, we develop a benchmark to evaluate hallucinations by large vision-language models (LVLMs) when
This paper proposes a benchmark to evaluate hallucinations by large vision-language models (LVLMs) when
generating free-form responses, specifically detailed descriptions,
based on a given image. 
% \zhizhong{We might need to shrink this claim: we ended up only doing image detailed descriptions. +``, specifically detailed descriptions, "}

The rapid advancement in large language models (LLMs)~\cite{zhao2023survey} has pushed 
% \yoni{propelled the development ...}
the development of large vision-language models
(LVLMs).~\cite{tsimpoukelli2021frozen,alayrac2022flamingo,li2022blip,li2023blip2,dai2023instructblip,huang2023kosmos,peng2023kosmos2,liu2023visual,ye2023mplug,zhu2023minigpt,chen2023minigpt2}
LVLMs take input text \emph{and images} and generate text responses
to enable multi-modal perception and comprehension.

LVLMs are largely built on LLMs and therefore inherit both their advantages and their disadvantages.
LLMs have been shown to produce hallucinations~\cite{ye2023cognitive,zhang2023siren}, 
generated text responses that are coherent and plausible but factually incorrect. LVLMs echo this behavior with generated text contradicting with the visual or text input~\cite{zhou2023analyzing}.
% generations which appear plausible but are factually incorrect, and
% current LVLMs also demonstrate this critical issue, models generate text contradicting evidence in the visual input~\cite{zhou2023analyzing}.
Hallucinations prevent the use of LVLMs in safety-critical situations and therefore evaluating and mitigating hallucinations in LVLMs is crucial for their deployment in such settings.
Determining the presence and cause of hallucinations in LVLMs remains an open question~\cite{wang2023evaluation,zhou2023analyzing}.

\begin{figure}[t]
    \centering
    \includegraphics[width=1.0\linewidth]{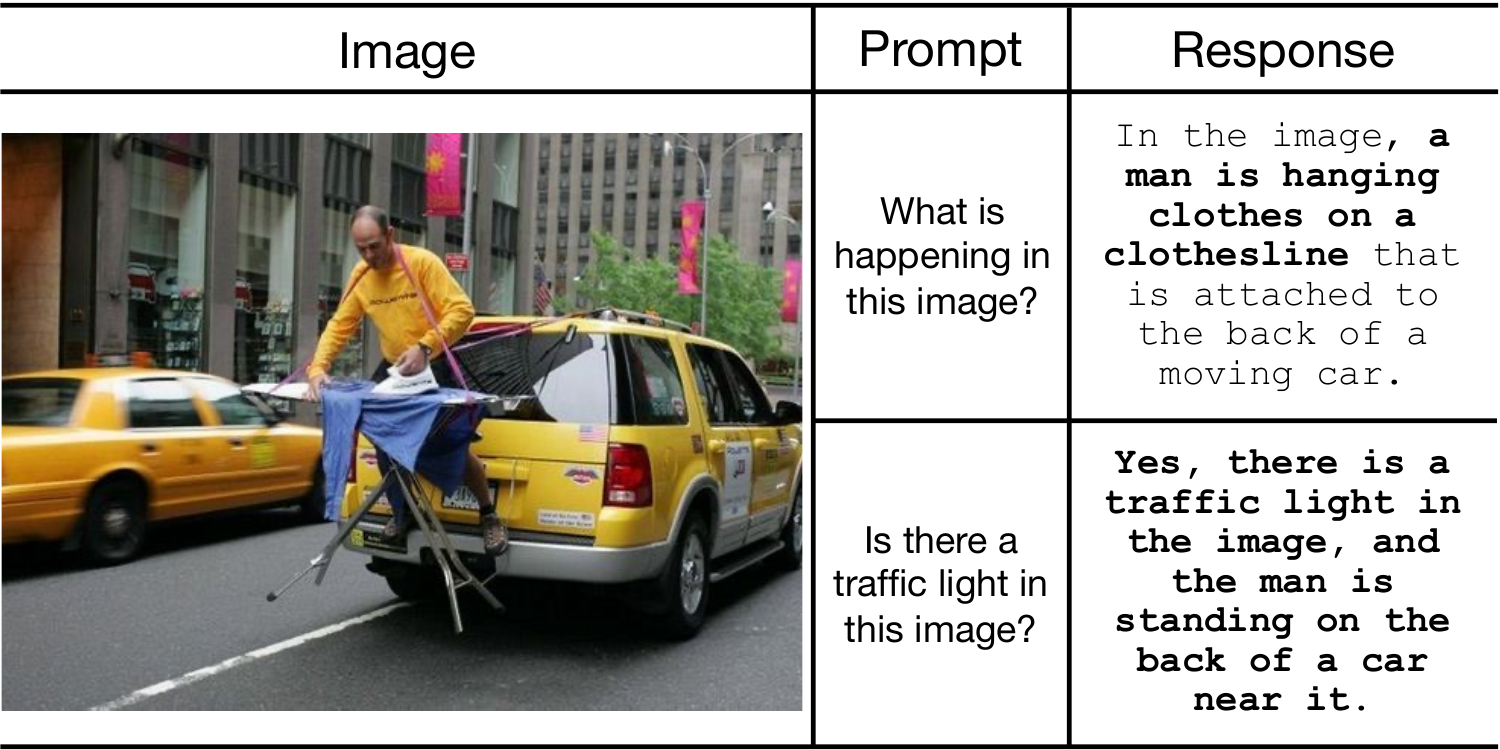}
    \caption{\textbf{Type I vs. Type II Hallucinations:} (Top) LVLMs prompted with concept-neutral instructions produce Type I hallucinations.
    (Bottom) Instructions specifying a concept produce Type II hallucinations. Examples from LLaVA-v1.5~\cite{liu2023improved}.}\label{fig:framework_hallucinations}
\vspace{-0.5cm}
\end{figure}

\begin{figure*}[t]
    \centering
    \includegraphics[width=1.0\linewidth]{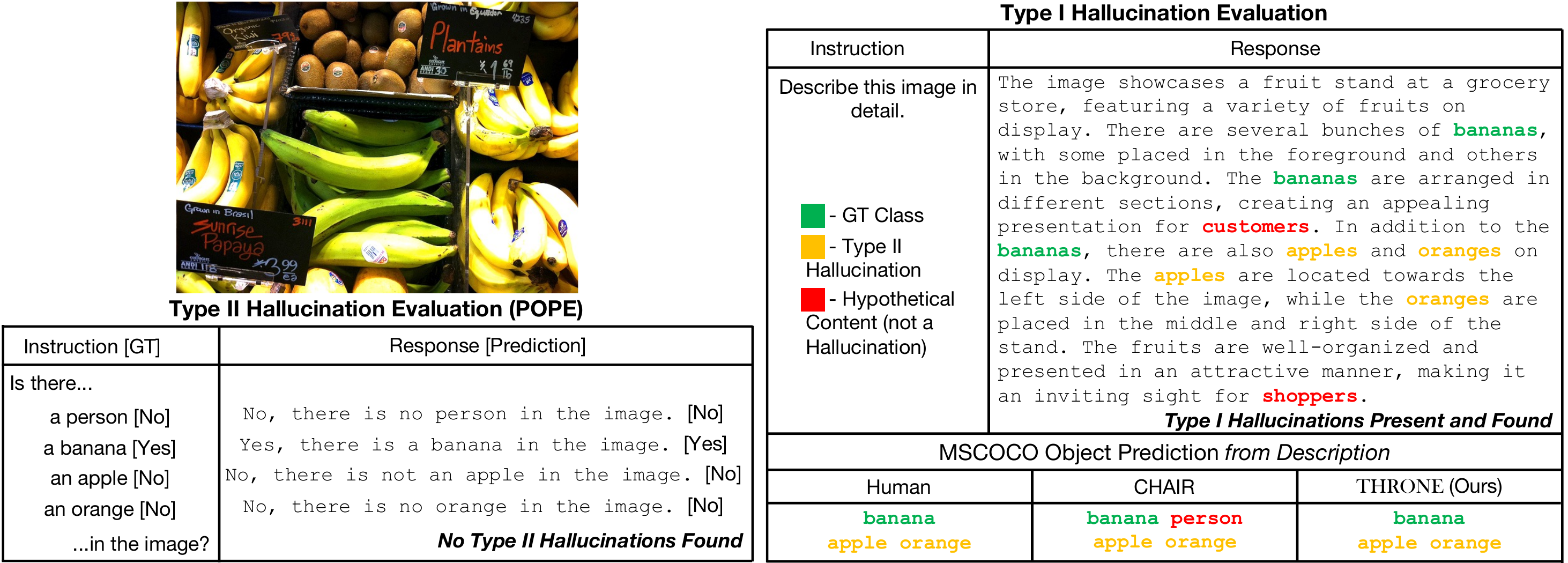}
    \caption{\textbf{A Comparison of POPE, CHAIR and \throne:} 
    \yoni{We illustrate different hallucination evaluation approaches with an example.}
    Directly querying LVLMs for object existence (person, banana etc.) using concept-specific instructions, as in POPE (bottom left),
    does not produce the same hallucinations as using concept-neutral instructions (right).
    We highlight the Type I hallucinations in {\color[HTML]{ffb800} orange}.
    CHAIR relies on exact text matching to a fixed set of objects and synonyms, thus incorrectly labels ``customers'' and ``shoppers'' as hallucinations, highlighted in {\color{red} red}.
    \throne is designed for the rich vocabulary and the free-form generations of modern LVLMs by harnessing LMs to establish object existence.
    By using an LM to pass judgement, our evaluation correctly captures ``customers'' and ``shoppers'' as hypothetical content in the free-form generation.}\label{fig:framework_overview}
    \vspace{-0.5cm}
\end{figure*}

We divide LVLM hallucinations into two types. 
Type I hallucinations occur in response to open-ended questions with a very large set of possible responses
---\eg~\texttts{What is happening in this image?}.
% Type I, occurring when an LVLM answers an open-ended question and \emph{is not} guided by a user's text instructions to consider a
% particular concept
Type II hallucinations are incorrect responses to a factual question regarding a specific concept about the image with a fixed set of options such as yes/no---\eg~\texttts{Is there a traffic light in this image?}
% Type II, occurring when an LVLM \emph{is} guided towards a specific concept by
% text instructions and invites a response from a fixed set of options like yes-no
\cref{fig:framework_hallucinations} illustrates the difference between these two types of hallucination.
Reducing hallucinations in both cases is required for useful, multi-purpose LVLMs.
However, later in \cref{fig:framework_overview} we observe that the same LVLM can give contradicting answers when being evaluated for Type I vs. Type II hallucinations.
This implies that measuring and reducing one type does not necessarily reduce the other.
% A LVLM which grounds \emph{all} generated responses to a given
% image---a necessary characteristic for accurate LVLMs---would
% not produce such contradictions. \zhizhong{$\leftarrow$ This sentence feels disconnected with the rest of the paper? Having such a model is not a point in our paper.}
% These contradictions show LVLMs do not ground generated
% responses to evidence present in a given image---a
% necessary characteristic for accurate LVLMs.

Existing works to evaluate LVLMs often avoid direct quantification of hallucinations
and instead develop comprehensive benchmarks that judge various other desirable abilities such as:
optical character recognition, fine-grained recognition and attribute detection~\cite{li2023seedbench,fu2023mme,liu2023mmbench}.
The extent of hallucinations in these benchmarks is obfuscated, since it is only one of many factors influencing other metrics. It requires % \zhizhong{... are not itemized...}
human effort to inspect individual predictions.
% Any hallucinations in answers to these benchmarks cannot be identified in these benchmarks without manual inspection.
% \zhizhong{I think these shortcomings can be merged into the second section, and only briefly summarized here.}
There are two major established works which specifically develop a benchmark for evaluating hallucinations in vision-language models: POPE~\cite{li2023evaluating} and CHAIR~\cite{rohrbach2018object},
which we discuss in detail in~\cref{sec:rw}.
However, we observe they both have shortcomings in effectively evaluating hallucinations:

% \yoni{Somewhat dangerous statement to make, as a reviewer might point their favorite recent submission on the topic }
POPE~\cite{li2023evaluating} is a recent work addressing Type II hallucinations with respect to object classes.
However, we find Type I and Type II hallucinations are disconnected, and that POPE gives an incomplete picture on LVLM hallucinations. 
Moreover, POPE systematically under-samples negative object categories leading to a large underestimation of Type II hallucinations (see~\cref{ssec:results_pope}).

CHAIR~\cite{rohrbach2018object} does address Type I hallucinations---establishing object category hallucination in short image captions
using simple text matching.
However, CHAIR is not suited to current LVLMs because the simple text matching it employs cannot comprehend abstract or hypothetical concepts present in today's LVLMs (see~\cref{fig:framework_overview}). Further, 
hand-crafted rules for each set of classes are required for usable text matching, and trivial model answers can attain a perfect CHAIR score.
% requires manual rules to ensure text matching is usable
% However POPE suffers from a number of important drawbacks:
% (1) The number of negative categories sampled per image for querying an LVLM is low (max $12$),
% which leaves many potential Type II hallucinations undiscovered---we experimentally establish this in~\ref{ssec:results_pope};
% (2) A key advantage of LVLMs is the generation of free-form responses which are not analyzed; and \zhizhong{This should be first and foremost; we should mention type I and type II are not correlated}
% (3) A relatively small number of images are evaluated ($\sim$$500$) which limits utility.

% CHAIR~\cite{rohrbach2018object} does address Type I hallucinations by establishing object hallucination in image captions
% using simple text matching.
% However, CHAIR was developed prior to the advent of current LVLMs and is unsuitable for Type I hallucinations for multiple
% reasons:
% (1) CHAIR utilizes simple text matching to establish positive object predictions in image captions---abstract concepts
% alluded to in LVLM responses cannot be comprehended correctly (see \cref{fig:framework_overview});
% (2) The metrics used in CHAIR do not measure the recall of ground-truth objects in the image caption, meaning a trivial model
% only making a single prediction gets a perfect score if it is correct;
% (3) CHAIR requires manual rules to be created to ensure synonyms of objects of interest sharing words do not lead to conflicting predictions.

To address the issues of current LVLM hallucination benchmarks,
we propose \throne (\emph{\textbf{T}ext-from-image \textbf{H}allucination \textbf{R}ecognition with \textbf{O}bject-probes for open-e\textbf{N}ded \textbf{E}valuation}).
\throne leverages language models (LMs) to evaluate Type I hallucinations in free-form, open-ended image descriptions with respect to a pre-defined object vocabulary of interest.
By utilizing LMs, \throne is able to accurately judge
whether an object mentioned in an LVLM response is implied to exist in the image or is abstractly mentioned with no implication about its existence (see~\cref{fig:framework_overview}).
% comprehend LVLM responses properly and correctly predict

Moreover, in \throne, we provide easy access to our benchmark, by leveraging open-source LMs that
can run on common GPUs, instead of relying on closed-source commercial models~\cite{openai2023gpt4} that are subject to arbitrary change, as done in other works~\cite{li2023seedbench,liu2023mmbench,bitton2023visitbench}.
Through combining multiple open-source LMs, we mitigate any single-model biases in judging hallucinations when calculating Type-I hallucination scores with \throne.
% Through a combination of multiple open-source LMs,
% we ensure \throne is robust to any biases of a particular LM in judging

We make four contributions:
\emph{first}, we establish an accurate and accessible benchmark to quantitatively evaluate object hallucinations in free-form responses,
leveraging LMs to judge the existence of Type I hallucinations---quantitatively showing half the judgement errors of CHAIR; % \yoni{what does that mean}\prannay{what about now? the judgement errors are halved}\@
\emph{second}, we evaluate a number of current LVLMs on \throne and demonstrate that
observed progress % \zhizhong{in model performance and reduction of...}
in reducing Type II hallucinations do not translate to a corresponding reduction
in Type I hallucinations;
\emph{third}, we show a recent method, POPE, is inaccurately capturing the extent of Type II hallucination discovery due to its sampling strategy; we mitigate this issue in our implementation of \throne
while presenting results for a complete version of POPE;\@ and % \zhizhong{TODO: if we also compare with CHAIR quantitatively, update this part}
\emph{fourth}, we provide a simple augmentation of visual instruction tuning data which significantly improves performance on Type I hallucinations
while maintaining or improving Type II hallucination performance.

\newcommand{\parb}[1]{\par{\noindent \bf #1}}
\cvprsection{Related Work}\label{sec:rw}
% \vspace{-0.2cm}
% \hao{this part is too long, need to cut down}
\parb{Hallucination Benchmarks for LVLMs.}
In response to the development of LVLMs, few evaluation benchmarks focusing on hallucinations have been introduced.
CHAIR~\cite{rohrbach2018object}, one of the first works to assess hallucinations, is designed for short \emph{image captions}.
CHAIR uses a fixed set of object classes (extended with their synonyms) as a set of text strings to find predicted object classes in image captions
via exact text matching.
Subsequently, each class matched in the caption is compared to the COCO~\cite{lin2014microsoft} captions and bounding box annotations to establish object hallucinations.
CHAIR was developed prior to current instruction-tuned LVLMs which produce long free-form responses ($10\times$ longer than image captions) with a diverse vocabulary,
limiting its applicability to modern LVLMs.
As shown in~\cref{fig:framework_hallucinations}, exact text matching in CHAIR
% cannot understand
% % abstract concepts alluded to in a free-form response, 
is prone to incorrectly matching the vocabiliary classes with abstract concepts in the free-form response and the 
and synonyms of the class list must be manually selected to prevent confusion during evaluation. For example, in CHAIR the word ``\texttts{chair}'' will match all responses for the phrase ``\texttts{toilet seat}'',
because the exact text matching classifies ``\texttts{seat}'' as the COCO class ``\texttts{chair}''.
Finally, CHAIR metrics compare the overall number of predicted objects to the overall number of predicted objects judged to be
hallucinations---ignoring the recall of ground-truth objects and the distribution of object classes. This means that a single correct prediction across the entire evaluation dataset along with a generic response otherwise
\eg~\texttts{A natural scene},
achieves a perfect score ($\frac{0\text{ hallucinated objects}}{1\text{ predicted objects}} = 0.0$).
See the Supplementary Material for a full overview of CHAIR.\@
In contrast, \throne uses pre-trained LMs
that go beyond direct synonym matching, 
% capable of comprehending abstract concepts,
 to automatically judge the existence of concepts and hallucinations in free-form responses.
In addition, our method considers both recall and precision to yield a holistic benchmark and does not require any manual curation of synonyms.
\throne and CHAIR both evaluate Type I hallucinations---hallucinations in response to concept-neutral prompts~\eg~\texttts{Describe this image in detail.}
% \yoni{Let's expand this point more regarding CHAIR's metrics, might be good to have a figure showing a CHAIR perfect score vs. \throne}
% \prannay{Can discuss, but I will give a full treatment of CHAIR in the SM}.

%
% has largely outgrown its usefulness \yoni{maybe be more specific when saying ``has largely outgrown it's usefulness'' e.g., As we present in Figure X, CHAIR's is not comprehensive enough for modern generations made by LVLMs}.
% Moreover, the exact word matching method cannot understand abstract concepts alluded to by free-form responses (see~\cref{fig:framework_hallucinations})
% and synonyms have to be manually cross-checked to prevent confusion during evaluation.
% Finally, CHAIR's metric compares the number of predicted objects to the number of predicted hallucinated objects (it does not measure the recall of ground-truth objects),
% meaning a single correct prediction across the entire evaluation dataset
% gives a perfect score ($\frac{0\text{ hallucinated objects}}{1\text{ predicted objects}} = 0.0$).

POPE~\cite{li2023evaluating} is a recently proposed benchmark to evaluate object hallucinations in LVLMs---specifically
Type II hallucinations, in which an LVLM is directly queried with a yes-no question regarding the existence of a particular object of the form:
\texttts{Is there \{a/an\} \{object\_class\_name\} in the image?}.
The LVLM response is parsed using simple rules to determine whether a Type II hallucination has occurred.
Precision and recall metrics are compiled using the parsed LVLM responses and the ground-truth annotation data.
Despite focusing on measuring hallucinations, POPE only queries an LVLM with 3 positive and 3 sampled negative questions per
COCO image~\ie~the evaluation is artificially balanced.
This means many potential hallucinations with respect to the COCO categories are not captured by their method.
In \cref{ssec:results_pope}, we show POPE dramatically underestimates Type II hallucinations and present the results of a complete version.
Our method, \throne, evaluates the prevalence of Type I hallucinations, which we observe are disjoint to Type II hallucinations.
% \subsection{Large Vision Language Models}\label{ssec:rw_lvlms}

\parb{Comprehensive Benchmarks for LVLMs} % \zhizhong{This is better as the second point because it is closer to our work.}
have recently grown in number.
MMBench~\cite{liu2023mmbench}
% MMHal-Bench~\cite{sun2023aligning},
and MM-Vet~\cite{yu2023mmvet}
assess various aspects of LVLM performance such as: color perception, celebrity recognition, and numerical calculation.
% \prannay{MMHal-Bench (from LLaVA-RLHF) are actually hallucination benchmarks but it is again arbitrary GPT-4 evaluation} \zhizhong{Are those concurrent work?} \prannay{First on arXiV: 25 Sep 2023, can remove}
However many of these works integrate evolving APIs which are modified often (or even discontinued) and are inherently stochastic. Over time this greatly reduces the consistency of these benchmarks. 
% However, these works use paid closed-sourced API models for evaluation.
% We believe this reduces access to the wider community and lacks reproducibility as closed-sourced API calls are stochastic and models are often updated or discontinued.
Exceptions are MME~\cite{fu2023mme} and SEED-Bench~\cite{li2023seedbench},
but the impact of Type II hallucinations on final metrics is conflated with a number of other aspects of model performance. % \zhizhong{I'm assuming they don't do Type-I hallucination?}
Our method, \throne, directly addresses Type I hallucinations, only making use of open-source language models and datasets.
% \yoni{How about renaming ````Comprehensive'' Benchmarks for LVLMs'' to ``LVLMs evaluation suites''}
% \prannay{I use ``comprehensive'' because the literature does, see abstracts for MMBench, SEED-Bench and paper title for MME, suites or something else is using a word not in the literature do not want to uncessarily confuse readers}
% Many of these works do some Type II hallucination evaluation, but as one facet of many other LVLM benchmarks informed our work.
% However, these works all use a closed-sourced paid-for OpenAI model (GPT-4 or ChatGPT) for evaluation.
% Exceptions are MME~\cite{fu2023mme} and SEED-Bench~\cite{li2023seedbench}, however these evaluation suites do not directly address
% hallucinations and any hallucination results are not easily extracted \yoni{What do you mean by not easily extracted? Is it that the metric is a mean/sum of a kitchen sink of evaluations so the final number is conflated? }.

\parb{Large Vision Language Models (LVLMs)} %\prannay{could have some cut down}
have rapidly developed by harnessing 
% on the back of 
advancements in large language models (LLM)~\cite{touvron2023llama,raffel2020,brown2020language} and by directly integrating pre-trained LLMs into their architectures.
In contrast to earlier vision-language models such as CLIP~\cite{radford2021learning,jia2021scaling},
LVLMs are generally comprised of a pretrained LLM and image encoder, aligned with a connector module of varying complexity.
Some works are highlighted here.
% We shall outline some recent LVLMs here.
% The rapid development of large language models (LLMs)\cite{touvron2023llama,raffel2020,brown2020language} in recent years has lead to an explosion
% in vision-langauge models which have the ability to process language instructions and visual inputs.
% Distinct to earlier vision-language models such as CLIP~\cite{radford2021learning}, ALIGN~\cite{jia2021scaling} and CoCa~\cite{yu2022coca},
% recent vision-language models are generally comprised of a pretrained LLM, a pretrained image encoder
% and an vision-language alignment module of varying complexity, from a linear projection to a transformer.
% In this paper, this type of vision-language model is termed a large vision-language model (LVLM).
% We shall give an overview of some recent LVLMs here, but given the growth in the field it is by no means exhaustive.
Frozen~\cite{tsimpoukelli2021frozen} is an early work fine-tuning the vision encoder to dynamically prefix the prompt to a frozen LLM.\@
Flamingo~\cite{alayrac2022flamingo} combines visual and language features using cross-attention layers in an otherwise frozen LLM.\@
BLIP-2~\cite{li2023blip2} uses a frozen image encoder to learn a Querying-Transformer (Q-Former) on image-text pairs that is used as the connector. This architecture is adapted in~\cite{dai2023instructblip} for dialogue via training with visual instruction tuning.
LLaVA~\cite{liu2023visual} uses COCO~\cite{lin2014microsoft} annotations and GPT-4~\cite{openai2023gpt4} to generate visual instruction
tuning data in a plain-text pipeline.
Combining this generated data with standard VQA datasets (VQAv2 etc.)
further boosts performance~\cite{liu2023improved}. Different works modify the training approach by using efficient adaptation~\cite{ye2023mplug,hu2022lora} multiple training stages~\cite{Qwen-VL}
or introduce the use of discrete tokens for localization~\cite{peng2023kosmos2}.
We note however, that most of these works evaluate performance on traditional vision-language datasets like VQAv2~\cite{antol2015vqa},
which do not consider the extent of hallucinations, a known problem with LLMs~\cite{zhang2023siren,ye2023cognitive}.

\cvprsection{\throne}\label{sec:framework}

\parb{Recap of existing methods.} POPE~\cite{li2023evaluating} and other benchmarks (MME~\cite{fu2023mme}, MMBench~\cite{liu2023mmbench}) directly query LVLMs with a restricted desired answer space,
\eg~yes-no (MME, POPE) and A-B-C-D multiple choice (MMBench), as shown in~\cref{fig:framework_pope}.
These benchmarks only consider such short answer formats,
whereas a key quality of LVLMs is in their ability to generate free-form coherent text.
Moreover, POPE, which addresses Type II hallucinations, under-samples negative classes meaning
hallucinations are dramatically underestimated (see~\cref{ssec:results_pope} and~\cref{fig:results_pope}).
In contrast, we skip class subsampling and \emph{enumerate all classes for every image}, ensuring a full evaluation of Type I hallucinations of the ground-truth classes.

CHAIR~\cite{rohrbach2018object} also evaluates Type I hallucinations,
but was developed when typical vision-language models could only generate short and simple captions similar to those in COCO Captions~\cite{fang2015captions}.
Moreover, it lacks accurate comprehension of responses (see the right side of~\cref{fig:framework_overview}) and ignores the recall of ground-truth objects.
In~\cref{ssec:results_ablations}, we describe quantitative evaluations, using a human oracle, which demonstrate \throne \emph{halves} the rate of
hallucination misjudgement in CHAIR.\@
See~\cref{sec:rw} and the Supplementary Material for more details on CHAIR and its shortcomings.\@
\cref{fig:framework_overview} shows an overview of the three aforementioned methods:
POPE, CHAIR and our method, \throne.

\cvprsubsection{Evaluating Hallucinations with \throne}\label{ssec:framework_ours}

% \zhizhong{I think (1-3) are the same point: they do short answers, we do open-ended answers.
% In that point, we can dive into the shortcomings of doing type II: LVLMs biased with Y/N, and LVLMs are often used for free-form answers.
% In addition, this can be merged into where we introduce type I vs II to save space and reduce redundancy}:
% (1) These benchmarks consider Type II hallucinations, if at all, ignoring Type I hallucinations;
% % (2) The chance of hallucination increases as response length increases\textbf{CITE LURE}, making it likely that these benchmarks, which lend themselves to ultra-short responses,
% % underestimate the extent of genuine hallucinations;
% (2) Some LVLMs exhibit extreme bias when answering yes-no questions~\cite{fu2023mme}---\eg~100\% ``yes'' answers on POPE,
% making it unclear whether hallucination is occurring or if this in artifact of the training data being simply extremely unbalanced; %, which can lead to genuine hallucinations being \emph{overestimated} in some cases,
% (3) Current LVLMs are capable of producing free-form responses containing image descriptions and explanations, which the aforementioned benchmarks do not test for hallucination; and
% (4) The number of negative object classes queried per image is very small ($3$ in POPE and $1$ in MME) meaning the full extent of hallucinations is not evaluated.

To address these limitations, we propose a framework, \throne, shown in~\cref{fig:framework_ours},
to evaluate the prevalence of Type I hallucinations
in LVLM responses conditioned on an image and a neutral text prompt.
% which occur when a response is largely conditioned on the image only, with the text prompt input short and neutral \ie~not focused on a specific class. \yoni{``is largely condiitoned'' is a bit unclear, ``occur'' and ``occurrences'' repeats...maybe saying something like ``To address these limitations, we propose \throne (shown in~\cref{fig:framework_ours}) which evaluates the occurrence of LVLM generations conditioned on the image with a neutral text prompt (Type I hallucinations).}

For each image in a labeled dataset, $\mathcal{I}$, addressing a set of classes, $\mathcal{C}$,
the LVLM is queried with the same instruction: \texttts{Describe this image in detail.}, regardless of image content.
The LVLM response, which is expected to be long free-form text containing an image description, is generated and stored.
Next, a publicly available, open-source, \emph{external} language model (LM) performs \emph{abstractive question answering} (AQA)
using the LVLM response as context and a question of the form: \texttts{Please answer yes or no. Is there \{a/an\} \{object class name\} in this image?} or similar,
\emph{for every class} in $\mathcal{C}$ (right side of~\cref{fig:framework_ours}).
By selecting an appropriate LM and using a simple prompt template (see \cref{sec:bench} for specific details),
we ensure the AQA response is either \texttts{yes} or \texttts{no}---our method does not require any additional parsing.
This is in contrast to other works which require added parsing or interpretation by a closed-source model.
% This is in contrast to POPE, MME, MMBench etc., which either use handcrafted rules to extract a
% prediction label \emph{or} use paid-for closed-source APIs (\eg~GPT-4) to extract a predicted label,
% limiting access and consistency---updating the model behind the API may change the predicted label and therefore the evaluated metric. \yoni{repeats the related work section}

After performing AQA on each response generated by the LVLM for every class in $\mathcal{C}$,
we obtain an array of predicted labels:
\vspace{-0.2cm}
\begin{align}\label{eq:framework_pred}
    \mathbf{\hat{Y}} \in {\{0,1\}}^{|\mathcal{I}|\times|\mathcal{C}|}
\vspace{-0.2cm}
\end{align}
where $0$/$1$ indicates a negative/positive existence \emph{judgement} by the LM with respect to the relevant LVLM response.

Similarly using the ground-truth data for $\mathcal{I}$, an array of ground-truth labels can be constructed:
\vspace{-0.2cm}
\begin{align}
    \mathbf{Y} \in {\{0,1\}}^{|\mathcal{I}|\times|\mathcal{C}|}
\vspace{-0.2cm}
\end{align}
% where $0$/$1$ indicates a negative/positive existence \emph{label} of a given class in the relevant image.

Using these two arrays, we calculate four metrics:
(1) Overall Precision, $P_{\textsc{ALL}}$;
(2) Overall Recall, $R_{\textsc{ALL}}$;
(3) Class-wise Precision, $P_{\textsc{CLS}}$;
and (4) Class-wise Recall, $R_{\textsc{CLS}}$.
Overall metrics are calculated in a class-agnostic manner.
Class-wise metrics are calculated in a class-conscious manner
by computing precision and recall for each category separately and then averaging.
This follows common practice in object detection and instance segmentation~\cite{lin2014microsoft,everingham10pascal}.
% \zhizhong{We should note right here that these are used to compute our main thing, which is $F^{0.5}_{CLS}$ with voting and all that. Now it just reads like we're proposing these four as \throne.} \zhizhong{We should also mention we don't do what POPE does and use the full negative set, and point to results showing why.}
% \begin{align}
%     P_{\textsc{ALL}} &= \frac{
%         \sum^{|\mathcal{I}|}_{i=1}\sum^{|\mathcal{C}|}_{j=1}Y_{i,j}\hat{Y}_{i,j}
%     }{\sum^{|\mathcal{I}|}_{i=1}\sum^{|\mathcal{C}|}_{j=1}\hat{Y}_{i,j}}\label{eq:framework_pall} \\
%     R_{\textsc{ALL}} &= \frac{
%         \sum^{|\mathcal{I}|}_{i=1}\sum^{|\mathcal{C}|}_{j=1}Y_{i,j}\hat{Y}_{i,j}
%     }{\sum^{|\mathcal{I}|}_{i=1}\sum^{|\mathcal{C}|}_{j=1}Y_{i,j}}\label{eq:framework_rall} \\
%     P_{\textsc{CLS}} &= \frac{1}{|\mathcal{C}|}\sum^{|\mathcal{C}|}_{j=1}\frac{
%         \sum^{|\mathcal{I}|}_{i=1}Y_{i,j}\hat{Y}_{i,j}
%     }{\sum^{|\mathcal{I}|}_{i=1}\hat{Y}_{i,j}} \label{eq:framework_pcls} \\
%     R_{\textsc{CLS}} &= \frac{1}{|\mathcal{C}|}\sum^{|\mathcal{C}|}_{j=1}\frac{
%         \sum^{|\mathcal{I}|}_{i=1}Y_{i,j}\hat{Y}_{i,j}
%     }{\sum^{|\mathcal{I}|}_{i=1}Y_{i,j}} \label{eq:framework_rcls}
% \end{align}
% \vspace{-0.2cm}

% \yoni{Not strong feeling, but can potentially move the definition of R, P to appendix, the intuitive description below might be more helpful}

% \zhizhong{I do think moving to appendix would be better. Most people should be familiar with these. Also our main thing is $F^{0.5}$ with voting and all that, so these are not really used.}

False positives in LVLMs reduce precision and are dominated by hallucinations---precision
indicates the extent of Type I hallucinations in LVLM responses.
The recall metrics inform the level of class coverage by an LVLM when producing image descriptions.
The class-wise metrics give a general measure of performance as the overall metrics
are skewed towards the most common categories.
A common way to combine precision, $P$, and recall, $R$, metrics is through the generalized F score, $F_\beta$:
\begin{equation}
    F^\beta = (1 + \beta^2)\cdot\frac{P\cdot R}{(\beta^2 \cdot P) + R}
\end{equation}
Prior work such as POPE~\cite{li2023evaluating},
use the common balanced F-score (or $F^1$-score) which equally weights precision and recall.
However, given that \throne is concerned with measuring hallucinations and is therefore
particularly interested in precision, we choose $\beta = 0.5$ or the $F^{0.5}$-score,
thus weighing precision twice as important as recall.
The $F^{0.5}$-score is commonly used in pandemic misinformation filters~\cite{misinfo2021},
recommender systems~\cite{grammar2014}, active stock selection~\cite{stock2020}
and other areas where false positives are costlier than false negatives.
From this we can calculate the overall and class-wise $F^{0.5}$-scores,
$F^{0.5}_\textsc{ALL}$ and $F^{0.5}_\textsc{CLS}$, respectively.
To mitigate against class imbalance issues, we use $F^{0.5}_\textsc{CLS}$ as the principle metric
of comparison between LVLMs in \throne.

\cvprsubsection{Ensuring Robustness via Ensembling}\label{ssec:frame_robust}

\begin{figure}[t]
    \centering
    \includegraphics[width=1.0\linewidth]{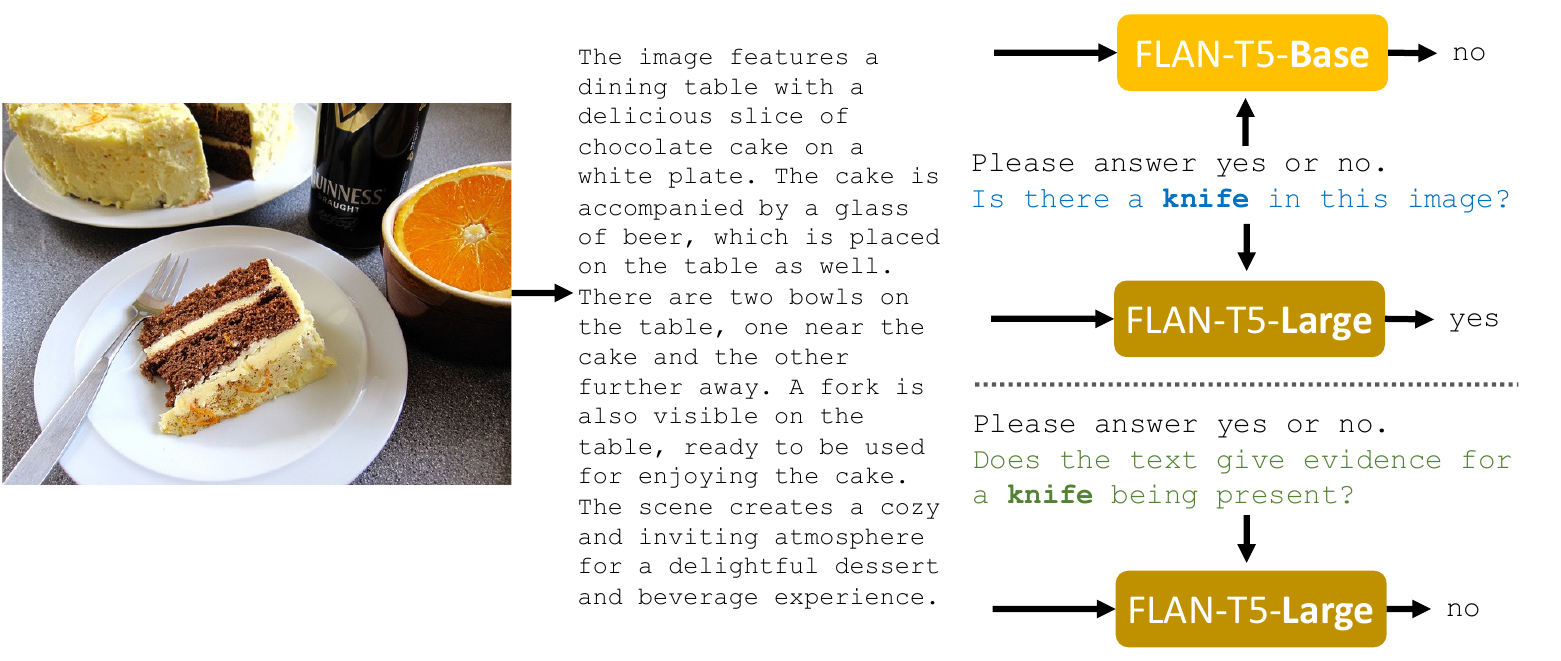}
    \vspace{-0.7cm}
    \caption{\textbf{AQA Ensembling in Evaluation:} Using different LMs or different prompts when running AQA on LVLM generated responses can produce opposing answers to \emph{identical prompts} or \emph{identical LMs}.
    To ensure \throne is robust to this, we ensemble \emph{multiple LMs} and \emph{multiple prompts} in our evaluation pipeline.}\label{fig:benchmark_prompt_model}
    %  \includegraphics[width=1.0\linewidth]{assets/figures/prompt_variation.pdf}
    % \caption{\textbf{Prompt QA Variation:} Different prompts with the same semantic meaning can produce opposing answers by \emph{the same LM}.
    % We ensemble \emph{multiple prompts} to ensure robustness.}
    % \label{fig:benchmark_model}
    % changed the combined label
    \vspace{-0.5cm}
\end{figure}

% Any LM used for AQA in \throne will  correct AQA comprehension for every (LVLM response, question) pair.\yoni{What is meant by exactly correct?}
Any LM used for AQA in \throne may misjudge Type II hallucinations---no LM is perfect.
\cref{fig:benchmark_prompt_model} (top) shows two different variants from the same model family (FLAN-T5~\cite{longpre23flan}),
yielding opposite responses when prompted with the same response and question.
Moreover, an LM may yield different answers to semantically identical questions,
despite conditioning on the same response---shown in~\cref{fig:benchmark_prompt_model} (bottom).
To ensure \throne is robust to spurious performance by any one LM,
we ensemble various LMs and semantically equivalent question formats.
The use of $N$ distinct LMs and $M$ distinct question formats yields $NM$ answers
for each (LVLM response, class) combination.
This set of $NM$ answers is combined based on equal voting by each (LM, question) pair
to ``elect'' the predicted answer.
Continuing the notation from~\cref{eq:framework_pred},
stacking predictions from each (LM, question) pair yields a 3D array of predicted labels:
$\mathbf{\bar{Y}} \in {\{0,1\}}^{|\mathcal{I}_{\text{OURS}}|\times|\mathcal{C}_{\textsc{OURS}}|\times NM}$.
To combine the answers from each (LM, question) pair via voting,
we require agreement between at least $k$ answers.
Where sufficient agreement does not exist we introduce an ``ignore'' label.
Mathematically, the elements in $\mathbf{\hat{Y}}$ (from \cref{eq:framework_pred}) are calculated as:
\vspace{-0.1cm}
\[
    \hat{Y}_{i,j} =
\begin{cases}
    0, & \sum_{k=1}^{NM} \bar{Y}_{i,j,k} \leq (NM - k) \\
    1, & \sum_{k=1}^{NM} \bar{Y}_{i,j,k} \geq k \\
    -1, & \text{otherwise}
\end{cases}
\vspace{-0.1cm}
\]
where an ``ignore'' label exists in $\mathbf{\bar{Y}}$,
it is removed from the calculations of $P_{\textsc{ALL}}$,
$R_{\textsc{ALL}}$, $P_{\textsc{CLS}}$, and
$R_{\textsc{CLS}}$,
as we cannot be confident in the AQA process for that particular (LVLM response, class) combination.
The choice of $k$ reflects the desired level of confidence.
We make use of a \emph{unanimous} voting mechanism ($k=NM$)---use
the prediction only if \emph{all} (LM, question) pairs agree, otherwise ignore.
% conduct experiments using three values of $k$
% \begin{itemize}
%     \item \emph{unanimous voting} ($k=9$)---use
% the prediction only if \emph{all} (LM, question) combinations agree, otherwise ignore.
%     \item \emph{simple majority} ($k=5$)---use
% the modal answer from the (LM, question) combinations, rendering ignoring impossible.
%     \item \emph{jury majority} ($k=8$)---use
% the prediction only if \emph{all} or \emph{all but one} of the (LM, question) combinations agree, otherwise ignore.
% \end{itemize}
Human evaluation of our benchmark and choice of voting mechanism is found in the Supplementary Material.
Once we have applied this voting mechanism, we can calculate the metrics
described at the end of~\cref{ssec:framework_ours}.

\cvprsection{Implementation}\label{sec:bench}
We provide details of our framework regarding the selection of public datasets, public LMs and LM prompts. 
% In this section, we provide specific details for \throne,
% using the general framework for evaluating Type I hallucinations described in the previous section.
% First we outline the dataset choice and the rationale behind this choice.
% Next, we describe the language models we use to perform AQA\@,
% making use of open-source models such that our benchmark remains easily accessible and consistent.
% Finally, we devise a way to ensure robustness in the AQA process and describe the methods
% to ensemble the outputs of multiple prompts from multiple language models when performing AQA on LVLM responses.

\cvprsubsection{Dataset}\label{ssec:bench_data}

Any proper evaluation of object hallucinations (a type of false positive error)
requires knowing, with certainty, which classes are absent in an image.
In our benchmark, we use COCO~\cite{lin2014microsoft} for a number of reasons:
(1) its annotations of 80 categories are exhaustive \emph{at an image level} (image-level recall$\approx$$99\%$~\cite{lin2014microsoft})---if
there are many \texttts{book} instances in a COCO image, at least one is annotated with bounding boxes;
(2) many LVLMs are partly trained on COCO data and so should be familiar with the set of categories;
(3) its images generally contain complex scenes suitable for generating long free-form descriptions unlike image
recognition datasets like ImageNet~\cite{Deng09}.

We utilize the validation set of COCO 2017,
which contains $|\mathcal{I}|=5000$ images and $|\mathcal{C}|=80$ categories.
Using the single LVLM text prompt \texttts{Describe this image in detail.}, we generate $5000$ responses.
As we query each LVLM response for each category in $\mathcal{C}$,
a single LM performs AQA $|\mathcal{I}| \times |\mathcal{C}| = 400$k times across the LVLM responses---one instance of
AQA per (image response, class) pair.
In~\cref{ssec:results_ours_obj365}, we also present results using Objects365~\cite{shao2019objects365} which is rarely used in LVLM training.

\cvprsubsection{Language Models}\label{ssec:bench_lms}
To assess Type I hallucinations in an LVLM response using \throne,
we require a language model (LM) which can answer questions on the
existence of object categories based on the LVLM response.
MMBench~\cite{liu2023mmbench} makes use of a ChatGPT model
to identify multiple choice answer selections and still reports mistakes.
In our experience, some LMs give rather incoherent judgements when used to assess hallucinations when the prompt is changed (see supplemental material).
Therefore for \throne, we choose FLAN-T5 models~\cite{longpre23flan,raffel2020}.
% specifically the \texttts{base}, \texttts{large} and \texttts{xl} variants. 
We make this choice because FLAN-T5 model family: 
(1) have undergone instruction tuning with thousands of tasks~\cite{longpre23flan}; 
(2) are open-source and therefore accessible to the community; 
(3) can fit locally on a single GPU for the models we consider; 
(4) follow user's instruction to only respond \texttts{yes} or \texttts{no}; and 
(5) are optimized for use in the free and public Text-Generation-Inference API~\cite{hf2023tgi} for acceleration.
% We make this choice for multiple reasons:
% (1) FLAN-T5 models are trained using FLAN instruction tuning data, which contains 1000s of tasks~\cite{longpre23flan};
% (2) The models are open-sourced and can be run locally, therefore accessible to the community, removing the need for a paid API in our benchmark;
% (3) The variants we choose have at most $3$ billion parameters and fit on commercial GPUs without the need for quantized inference;
% (4) When prompted correctly, these models faithfully produce responses of either \texttts{yes} or \texttts{no} only,
% removing the need for extended parsing;
% (5) The models are optimised for use in the free and public Text-Generation-Inference API~\cite{hf2023tgi},
% which accelerates evaluation should \throne be used on larger datasets in future.
As described in~\cref{ssec:frame_robust}, we utilize $N$ LMs to ensure our method is robust.
Specifically, we use $N=3$ variants of FLAN-T5, namely:
%  \yoni{can enumerate the FLAN models and parameters in-line to save space}\prannay{lets keep for now matches the enumeration of prompts}
FLAN-T5-Base ($250$M parameters), FLAN-T5-Large ($780$M parameters), and FLAN-T5-XL ($3$B parameters).

\vspace{0.2cm}
\cvprsubsection{Prompt Ensembling}\label{ssec:bench_prompt}
To guarantee each of these FLAN-T5 variants
faithfully produce responses of either \texttts{yes} or \texttts{no} only during AQA,
we use the following input template to each LM, reflecting the format used when
training FLAN-T5\footnote{\href{https://tinyurl.com/5n6nexze}{\texttt{https://tinyurl.com/5n6nexze}}}:
\begin{quotation}
    % \vspace{-0.1cm}
    \scriptsize
    \noindent\texttts{Text: \emph{\{LVLM Response\}} 
    Read the text about an image and answer the question.
    Question: Please answer yes or no.\ \emph{\{Question\}}}
    % \vspace{-0.1cm}
\end{quotation}

We use $M=3$ semantically identical questions:
\begin{itemize}
    \scriptsize
    \item \texttts{Is there a/an \{\emph{class\_name}\} in this image?}
    \item \texttts{Does the text imply a/an \{\emph{class\_name}\} is in the image?}
    \item \texttts{Does the text explicitly mention a/an \{\emph{class\_name}\} is in the image?}
\end{itemize}

As outlined in~\cref{ssec:frame_robust}, we use a \emph{unanimous voting} mechanism to combine the answers
from each (LM, question) pair and so $k = 3\times 3 = 9$.

\begin{table}
\centering
\begin{adjustbox}{width=\linewidth}
\begin{tabular}{l|c|cccc|ccc>{\columncolor[HTML]{EFEFEF}}c}
\toprule
% \scriptsize
% \setlength\fboxsep{-10.0pt}
Model                                         & $L$ & $P_\textsc{ALL}$                     & $R_\textsc{ALL}$                     & $F^1_{\textsc{ALL}}$                 & $F^{0.5}_{\textsc{ALL}}$             & $P_\textsc{CLS}$                     & $R_\textsc{CLS}$                     & $F^1_{\textsc{CLS}}$                 & $F^{0.5}_{\textsc{CLS}}$             \\ \midrule
Adapter-v2~\cite{gao2023llamaadapter}   & 514 & 63.6                                 & 73.3                                 & 68.1                                 & 65.3                                 & 68.2                                 & 70.6                                 & 69.4                                 & 68.7                                 \\
Adapter-v2.1~\cite{gao2023llamaadapter} & 512 & 63.8                                 & 73.7                                 & 68.4                                 & 65.5                                 & 67.4                                 & 71.2                                 & 69.3                                 & 68.1                                 \\
InstructBLIP~\cite{dai2023instructblip}       & 525 & 70.8                                 & {\color[HTML]{FF0000} \textit{74.3}} & {\color[HTML]{FF0000} \textit{72.5}} & 71.5                                 & 77.2                                 & {\color[HTML]{FF0000} \textit{71.9}} & {\color[HTML]{FF0000} \textit{74.5}} & 76.1                                 \\
Otter-Image~\cite{li2023otter}                & 257 & 33.0                                 & 31.2                                 & 32.1                                 & 32.7                                 & 25.2                                 & 16.9                                 & 20.2                                 & 22.9                                 \\
MiniGPT4~\cite{zhu2023minigpt}                & 473 & \textit{81.7} & 59.8                                 & 69.0                                 & 76.1                                 & {\color[HTML]{FF0000} \textit{79.9}} & 61.8                                 & 69.7                                 & 75.5                                 \\
MiniGPT-v2~\cite{chen2023minigpt2}            & 381 & 79.0                                 & 66.6                                 & 72.3                                 & 76.2 & 77.6                                 & 67.0                                 & 71.9                                 & 75.2                                 \\
mPLUG-Owl~\cite{ye2023mplug}                  & 555 & 55.5                                 & 71.9                                 & 62.6                                 & 58.1                                 & 66.3                                 & 68.3                                 & 67.3                                 & 66.7                                 \\
LRV-Instruction-v2~\cite{liu2023aligning}     & 103 & {\color[HTML]{FF0000} \textit{82.0}} & 56.7                                 & 67.0                                 & 75.3                                 & {\color[HTML]{FF0000} \textit{78.4}} & 58.8                                 & 67.2                                 & 73.5                                 \\
LLaVA-v1.3*~\cite{liu2023visual}              & 532 & 80.5                                 & 65.2                                 & 72.1                                 & 76.9 & {\color[HTML]{FF0000} \textit{79.9}} & 65.3                                 & 71.9                                 & {\color[HTML]{FF0000} \textit{76.5}} \\
LLaVA-v1.5~\cite{liu2023improved}             & 509 & 68.1                                 & 61.0                                 & 64.4                                 & 66.6                                 & 69.9                                 & 56.4                                 & 62.5                                 & 66.8                                 \\ 
LLaVA-Mistral~\cite{jiang2023mistral,liu2024llavanext}                   & 524 & {\color[HTML]{0432FF} \textbf{86.8}}                                 & {\color[HTML]{0432FF} \textbf{71.8}} & {\color[HTML]{0432FF} \textbf{78.3}} & {\color[HTML]{0432FF} \textbf{83.6}}                                 & {\color[HTML]{0432FF} \textbf{84.4}}                                     & {\color[HTML]{0432FF} \textbf{64.2}} & {\color[HTML]{0432FF} \textbf{70.8}} & {\color[HTML]{0432FF} \textbf{77.5}} \\
\bottomrule
\end{tabular}
\end{adjustbox}
\caption{\textbf{\throne Results with COCO} for a selection of instruction-tuned LVLMs.
We select $F^{0.5}_{\textsc{CLS}}$ as the principal metric for evaluation in our benchmark to balance across classes and to prioritize precision (which reflects the extent of hallucination) over recall.
Best and second-best performance are denoted by {\color[HTML]{0432FF} \textbf{blue}} and {\color[HTML]{FF0000} \textit{red}}, respectively.
*Our implementation using official code to enable fair comparison. $L$ corresponds to the median response length (measured in  $\#$ of characters).
}\label{tab:results_main}
\end{table}

% \begin{table*}[t]
% \centering
% \scriptsize
% \setlength\tabcolsep{3pt}

% \end{table*}

\cvprsection{Evaluation Results}\label{sec:results}
In this section, we:
(1) outline our LVLM selection and reasoning;
(2) present \throne on COCO for evaluating Type I hallucinations;
(3) extend \throne to Objects365 (containing a larger vocabulary); 
(4) analyze and extend POPE to enable improved evaluation of Type II hallucinations; and
(5) highlight results from our ablation studies found in the Supplemental Material. %\zhizhong{Merge improvement over POPE into ablation}

% In this section, first, we outline our method for LVLM selection.
% Second, we provide LVLM results on \throne using COCO as discussed in~\cref{sec:bench}.
% Third, we extend our evaluation to Objects365 which has a larger object vocabulary and is not contained in the training data of any evaluated LVLM.\@
% Next, we discuss the show limitations of POPE due to undersampling by comparing results with a complete version.
% %  the undersampling of negatives in POPE also effects results on \throne.
% Finally, we provide headlines for ablation experiments conducted with full details in the Supplementary Material. 
% \zhizhong{Can we say we ablate our sampling strategy?}
% \prannay{Table Added}
% \prannay{could delete this sentence} Moreover, this comparison demonstrates performance on our benchmark, which assesses Type I hallucinations,
% is orthogonal to performance on POPE, which assesses Type II hallucinations.
% \prannay{Will have to move to supplementary Finally, we will confirm the robustness of our benchmark using additional experiments.}
% section/subsection headings may need redoing}

\cvprsubsection{Models}
For fair comparison between existing LVLMs,
each publicly available model we evaluate uses an LLM with $\sim$$7$B parameters.
The LVLMs generally have different sized image encoders--\eg~LLaVA~\cite{liu2023visual} uses
a CLIP ViT-L/14~\cite{dosovitskiy2020image,radford2021learning} with an input resolution of $336\times336$,
while InstructBLIP~\cite{dai2023instructblip} uses a ViT-g/14~\cite{zhai2022scaling} trained with EVA~\cite{fang2023eva} and
an input resolution of $224\times224$. Note that image encoder size and resolution is not something we can easily control in a pre-trained model.
Each model we consider contains instruction tuning in the final training phase---instruction
tuned models provide free-form descriptions; \throne focuses on models that \emph{can} generate free-form descriptions.
\Eg~we leave out BLIP-2~\cite{li2023blip2} (response median length 31 characters) in favor of
% \redact{\Eg~BLIP-2~\cite{li2023blip2} produces ``detailed descriptions'' which have a median response length of 31 characters,
% resembling captions in the COCO Caption dataset, which are often incoherent and lack detail---whereas
% InstructBLIP produces responses with improved style and detail with a median response length of 525.}
InstructBLIP (median response length 525).
% The Supplementary Material contains comprehensive details on each of the LVLMs that we evaluate.
% Moreover, models trained without instruction tuning generate simple captions similar to those
% in the COCO Caption dataset, which are often incoherent and lack detail.

\cvprsubsection{\throne Results on COCO}\label{ssec:results_ours}

Results are shown in \cref{tab:results_main}.
The principal metric that we use to judge model performance is the classwise $F^{0.5}$-score ({\fboxsep=0.5pt\colorbox[HTML]{EFEFEF}{highlighted gray}}).
We also report all the metrics outlined in \cref{ssec:framework_ours}, ($P,~R, F^1, F^{0.5}$) for
 overall (left) and class-wise averaging (right), utilizing the \emph{unanimous} voting presented in \cref{ssec:frame_robust}.
See the Supplementary Material for results and an analysis of different voting mechanisms.
% \redact{
% We observe that Qwen-VL performs best on our benchmark, with LLaVA-v1.3 performing second best.
% Interestingly, we find the successor model, LLaVA-v1.5, performs markedly worse than its predecessor.
% Similarly, despite LLaMA-Adapter-v2.1 and MiniGPT-v2 gaining a lot of ground in POPE (see~\cref{fig:results_pope}) compared to LLaMA-Adapter and MiniGPT4,
% both model perform slightly worse on \throne.}
These results demonstrate that improvements on other benchmarks (POPE, MME, MMBench etc.) may be orthogonal and potentially at odds with improved performance on
\throne.
Using the results of the 11 LVLMs that we evaluate, \throne and POPE, which measure Type I and Type II hallucinations, respectively,
have a Spearman's rank correlation coefficient of just $0.2$, and \throne vs POPE-C has just $0.4$---the relationship between performance on POPE and \throne \emph{on the same dataset} is far from monotonic. 
% \zhizhong{Do we have spearman's correlation between POPE ranking and \throne ranking?}
% \prannay{Maybe remove, why bring up MMBench}
% Moreover, Otter-Image, which performs well on MMBench, performs poorly on our benchmark demonstrating it is prone to generating Type I
% hallucinations.
For class-wise precision---$P_{\textsc{CLS}}$, the best performing models hallucinate $\sim$20\% of the objects. We show in the Supplementary Material that the
vast majority of false positive objects in the free-form image descriptions evaluated are direct hallucinations rather than misclassifications of visually similar objects
(\eg~mistaking a squash racket for a tennis racket).  % \yoni{What is the difference between the two in this context?}
% (see the Supplementary Material for manually labeled quantitative evidence for this claim).
These results demonstrate that much work still remains to adequately suppress Type I hallucinations in LVLMs.

\begin{table}[ht]
\centering
\scriptsize
\begin{adjustbox}{width=\linewidth}
\setlength\tabcolsep{2pt}
\begin{tabular}{@{}l|rrrr|rrr>{\columncolor[HTML]{EFEFEF}}r}
\toprule
Model                                         & $P_\textsc{ALL}$                     & $R_\textsc{ALL}$                     & $F^1_{\textsc{ALL}}$                 & $F^{0.5}_{\textsc{ALL}}$             & $P_\textsc{CLS}$                     & $R_\textsc{CLS}$                     & $F^1_{\textsc{CLS}}$                 & $F^{0.5}_{\textsc{CLS}}$             \\ \midrule
Adapter-v2~\cite{gao2023llamaadapter}   & 46.7                                 & 33.9                                 & 39.3                                 & 43.4                                 & 48.9                                 & 28.5                                 & 36.0                                 & 42.8                                 \\
Adapter-v2.1~\cite{gao2023llamaadapter} & 46.8                                 & 34.0                                 & 39.4                                 & 43.5                                 & 48.8                                 & 28.8                                 & 36.2                                 & 42.8                                 \\
InstructBLIP~\cite{dai2023instructblip}       & 54.5                                 & 37.2                                 & 44.2                                 & 49.8                                 & 53.7                                 & 33.6                                 & 41.3                                 & 48.0                                 \\
Otter-Image~\cite{li2023otter}                & 21.4                                 & 12.7                                 & 16.0                                 & 18.8                                 & 9.5                                  & 4.4                                  & 6.0                                  & 7.7                                  \\
MiniGPT4~\cite{zhu2023minigpt}                & 53.0                                 & 32.9                                 & 40.6                                 & 47.2                                 & 49.9                                 & 31.9                                 & 39.0                                 & 44.9                                 \\
MiniGPT-v2~\cite{chen2023minigpt2}            & 54.5                                 & 36.0                                 & 43.4                                 & 49.4                                 & 51.3                                 & {\color[HTML]{FF0000} \textit{34.6}} & 41.3                                 & 46.8                                 \\
% Qwen-VL-Chat~\cite{Qwen-VL}                   & {\color[HTML]{0432FF} \textbf{58.3}} & {\color[HTML]{FF0000} \textit{39.1}} & {\color[HTML]{0432FF} \textbf{46.9}} & {\color[HTML]{0432FF} \textbf{53.1}} & {\color[HTML]{0432FF} \textbf{57.8}} & {\color[HTML]{0432FF} \textbf{35.9}} & {\color[HTML]{0432FF} \textbf{44.3}} & {\color[HTML]{0432FF} \textbf{51.5}} \\
mPLUG-Owl~\cite{ye2023mplug}                  & 43.7                                 & 33.4                                 & 37.8                                 & 41.2                                 & 48.2                                 & 29.0                                 & 36.2                                 & 42.6                                 \\
LRV-Instruction-v2~\cite{liu2023aligning}     & 57.5                                 & 26.7                                 & 36.5                                 & 46.7                                 & 51.4                                 & 26.6                                 & 35.1                                 & 43.3                                 \\
LLaVA-v1.3*~\cite{liu2023visual}              & {\color[HTML]{FF0000} \textit{57.6}} & 32.9                                 & 41.9                                 & 50.1                                 & 52.6                                 & 30.5                                 & 38.6                                 & 45.9                                 \\
LLaVA-v1.5~\cite{liu2023improved}             & 54.0                                 & {\color[HTML]{0432FF} \textbf{39.5}} & {\color[HTML]{FF0000} \textit{45.6}} & {\color[HTML]{FF0000} \textit{50.3}} & {\color[HTML]{FF0000} \textit{53.9}} & 34.3                                 & {\color[HTML]{FF0000} \textit{41.9}} & {\color[HTML]{FF0000} \textit{48.4}} \\ 
LLaVA-Mistral~\cite{jiang2023mistral,liu2024llavanext}                   & {\color[HTML]{0432FF} \textbf{58.3}} & {\color[HTML]{FF0000} \textit{39.1}} & {\color[HTML]{0432FF} \textbf{46.9}} & {\color[HTML]{0432FF} \textbf{53.1}} & {\color[HTML]{0432FF} \textbf{57.8}} & {\color[HTML]{0432FF} \textbf{35.9}} & {\color[HTML]{0432FF} \textbf{44.3}} & {\color[HTML]{0432FF} \textbf{51.5}} \\\bottomrule
\end{tabular}
\end{adjustbox}
\caption{\textbf{\throne Evaluation with Objects365.} Evaluation results for a selection of instruction-tuned LVLMs, we use a subset of Objects365 for the \throne 
 evaluation.
Best and second-best performance are denoted by {\color[HTML]{0432FF} \textbf{blue}} and {\color[HTML]{FF0000} \textit{red}}, respectively.
*Our implementation using official code to enable fair comparison.
}\label{tab:results_obj365}
\vspace{-0.4cm}
\end{table}

\cvprsubsection{\throne Results on Objects365}\label{ssec:results_ours_obj365}

\begin{figure}[h!]
    \centering
    \includegraphics[width=1.0\linewidth]{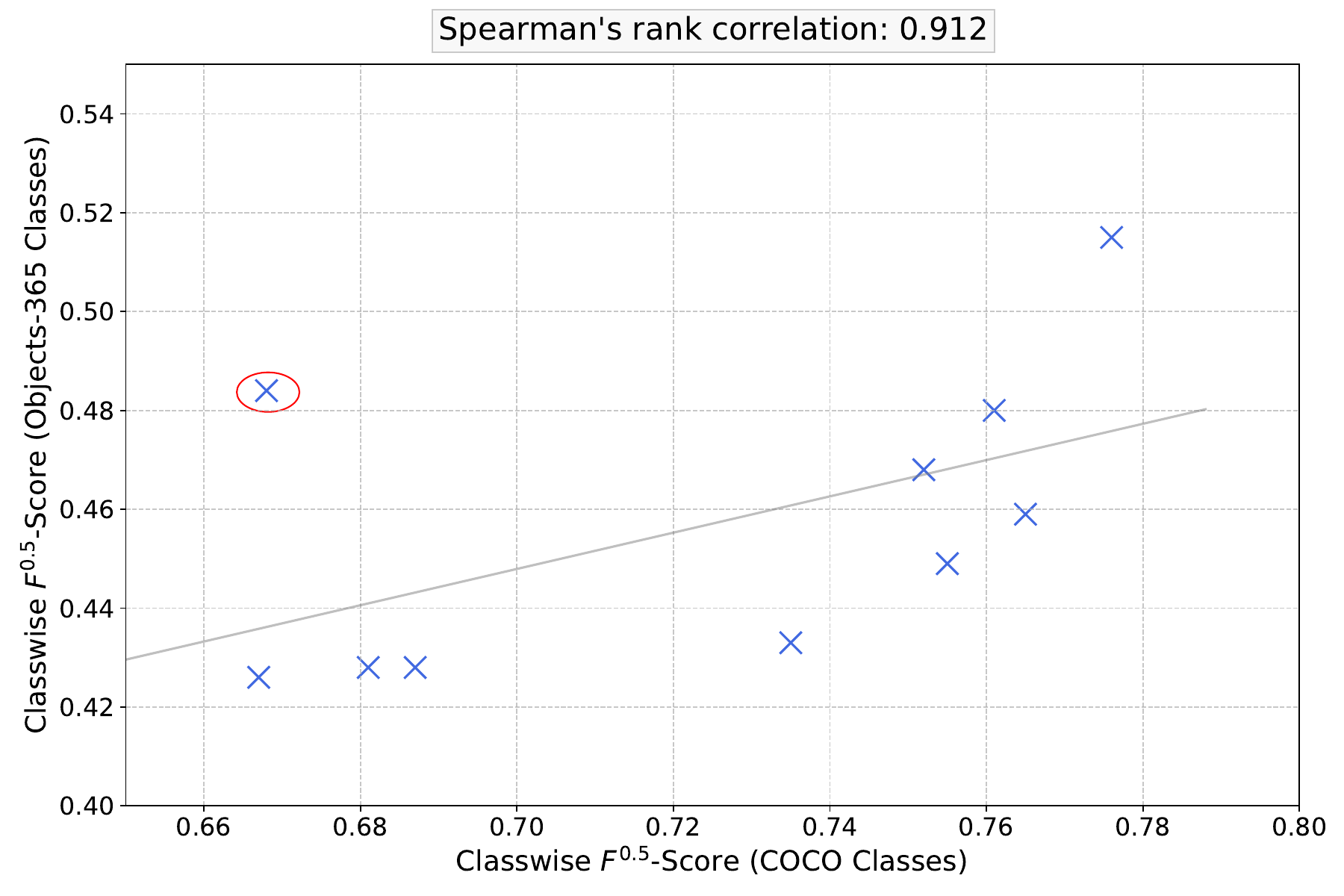}
    \vspace{-0.8cm}
    % throne scores for COCO and Objects365. We present a comparision of the performance of the benchmakred LVLMs when using different underlying datasets with different scales and number of categories. We observe that despite drastic variations in the data, throne scores generalize and and rank-consistent with a spearman's correlation of r=0.900.
    \caption{\textbf{Comparing \throne on COCO and Objects365.}
    We observe that despite variations in the data distribution, \throne metrics,
    which measure Type I hallucinations generalize and have a strong Spearman's rank correlation of $r=0.900$.
    One model (red circle) designated as an outlier and ignored when calculating ranking correlation.
    % \prannay{Could use this one or the one which splits O365 in COCO and Non-COCO}
    }\label{fig:coco_obj365_corr}
    \vspace{-0.5cm}
    \end{figure}

Many LVLMs train on COCO directly or indirectly, thus to demonstrate generality
% To demonstrate generality beyond COCO, which many LVLMs train on directly or indirectly, % \yoni{<-re-write this sentence maybe?}\zhizhong{Is the blurb about InstructBLIP and LLaVA exhaustive?} 
% There may be concern that \throne, which uses COCO for evaluation will favor models which have trained on COCO for captioning tasks
% (\eg InstructBLIP~\cite{dai2023instructblip} uses COCO captions as training data and LLaVA~\cite{liu2023visual,liu2023improved}
% uses COCO captions and bounding box annotations to generate visual instruction tuning data).
% To alleviate these concerns, 
we apply \throne to the Objects365 dataset~\cite{shao2019objects365}.
Like COCO, Objects365 aims to be exhaustive in its image-level class labeling (it aims to label at least one instance for each class present), but it has a larger object vocabulary and is not used as training data for the LVLMs that we evaluate.
To gather a manageable subset of the Objects365 validation set ($80$k images),
we use the natural sampling algorithm from~\cite{lee2022rethinking},
resulting in $5110$ images (the COCO validation set has $5$k images).
We present the results for \throne on Objects365 in~\cref{tab:results_obj365}.
% \redact{
% We observe again that Qwen-VL-Chat~\cite{Qwen-VL} performs best on \throne.
% Unlike for COCO (\cref{tab:results_main}), we find that LLaVA-v1.5~\cite{liu2023improved} performs better than LLaVA-v1.3~\cite{liu2023visual} on Objects365,
% but note this is largely due to increased recall.
% This likely results from additional training data exposing LLaVA-v1.5 to a greater number of concepts than LLaVA-v1.3,
% which only trains on visual instruction tuning data derived from COCO.
% Similarly, LLaMA-Adapter-v2.1 and MiniGPT-v2 marginally improve over LLaMA-Adapter and MiniGPT4, respectively, by having better recall, despite showing much larger gains on POPE.}
\Cref{fig:coco_obj365_corr} shows the strong correlation in \throne performance between evaluating on COCO and Objects365.
This demonstrates that measuring Type I hallucinations in an LVLM using \throne with a relatively small dataset like COCO,
is indeed indicative of the intrinsic level of Type I hallucination in a given LVLM.

\cvprsubsection{Completing POPE for Type II Hallucinations}\label{ssec:results_pope} % \zhizhong{Frame this as differences from type I or ablation or something. At least say "Improvements over POPE" rather than limitations}
\begin{figure}
\centering

\includegraphics[width=\linewidth]{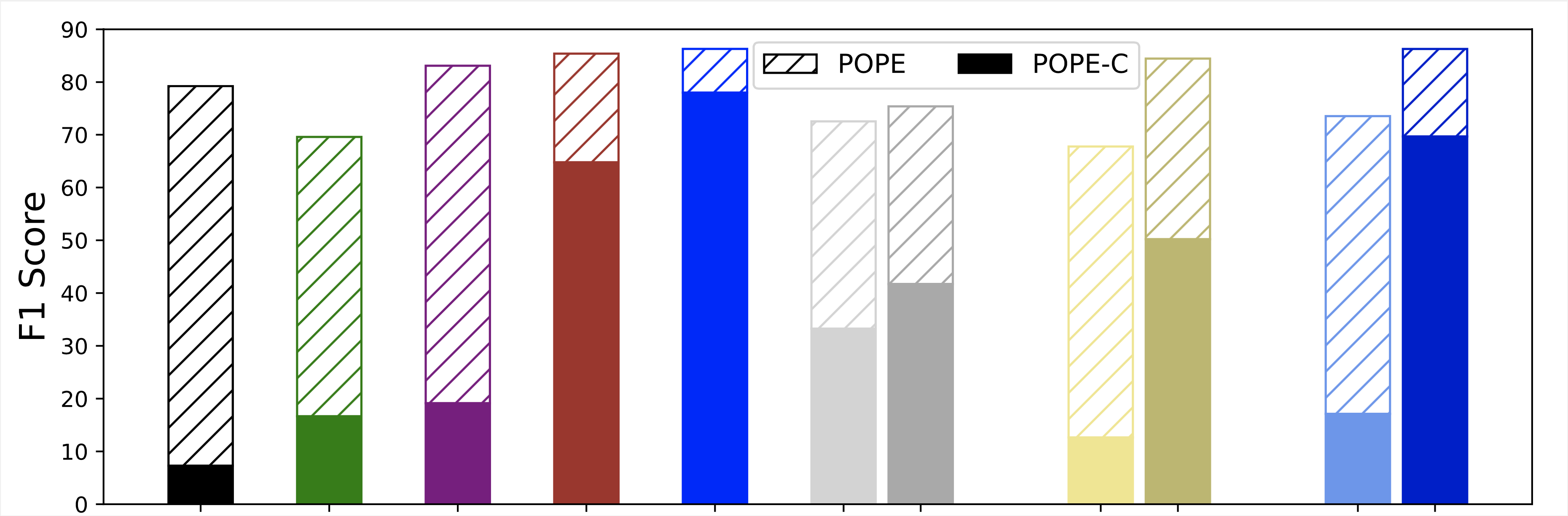}
% \vspace{-0.9cm}
\caption{\textbf{Instability of POPE to complete evaluation of Type II Hallucinations.} Extending POPE to an exhaustive analysis on all COCO images and classes (POPE-C) leads to a dramatic reduction in performance across all 11 models.
POPE is sensitive to the sampling mechanism, and by undersampling negative classes, POPE severely underestimates Type II hallucinations in LVLMs.}\label{fig:results_pope}
% Instability of POPE to complete evaluation of Type II Hallucinations.  
% \caption{POPE does not query every category of interest for Type II hallucinations.
% Extending POPE to an exhaustive analysis on all COCO leads to a dramatic reduction in performance across all models,
% demonstrating the fragility of POPE to the sampling mechanism. By under-sampling negative classes, POPE severely underestimates Type II hallucinations in LVLMs.
% }
\vspace{-0.5cm}
\end{figure}

Our experiments have used \throne to evaluate the prevalence of Type I hallucinations in LVLM responses on COCO.
POPE evaluates Type II hallucinations on COCO, but we find POPE is largely incomplete.
First, POPE only evaluates on 500 COCO validation set images.
Second, for each image only a subset of classes (at most 12) are evaluated---each image is only queried with 15\% of %\yoni{the}
possible questions.
Finally, POPE artificially balances evaluation questions between positives and negatives, despite object class existence in images being inherently imbalanced.  
The above reasons make the evaluation of Type II hallucinations using POPE insufficient (see \cref{sec:rw} for more details on POPE). % \yoni{maybe use a different word, e.g. ``insufficient''}\prannay{I do not think it is insufficient I think it is wrong.} \zhizhong{Let's not point fingers at other papers with blanket statements. All we did is making it complete, so we can just say POPE is very incomplete.}.
To correct this, we complete POPE by using all images to exhaustively query each LVLM \emph{for every class} in the COCO vocabulary, as done in \throne.
We name this POPE-Complete (POPE-C).
\cref{fig:results_pope} shows the extreme difference in evaluation between POPE and our exhaustive version, POPE-C---reporting
$F^1$-score (POPE does not utilize $F^{0.5}$-score).
As POPE only evaluates at most $9$ negative classes, only a small subset of potential hallucinations of COCO classes are evaluated,
thereby heavily underestimating the extent of Type II hallucination.
For each LVLM analyzed, we observe a large---in many cases an extreme---reduction in precision and therefore in F1-score.
% \redact{\Eg Otter-Image, LRV-Instruction-v2, LLaVA-v1.3, MiniGPT4 and mPLUG-Owl all have an F1-score reduction of at least 50\%}.

Our evaluation contains three pairs of LVLMs in which one is the follow-up work to the other, 
% \redact{(MiniGPT4$\rightarrow$MiniGPT-v2, LLaMA-Adapter-v2$\rightarrow$v2.1, LLaVA-v1.3$\rightarrow$LLaVA-v1.5),}
where each follow-up work generally trains on more data for more tasks with a more advanced language model.
Comparing the right hand side of~\cref{fig:results_pope} to \cref{tab:results_main},
we observe that these follow-up works generally show an improvement in POPE (and POPE-C),
but surprisingly indicate a small reduction in performance on \throne with COCO.
This observation suggests that progress in reducing Type II hallucinations can be orthogonal
to reducing Type I hallucinations.

\cvprsubsection{Ablations}\label{ssec:results_ablations}
In the Supplementary Material we present three key ablation experiments and give an executive summary of results here.

First, after subsampling COCO images and LVLM responses, we replace the LMs in \throne
with human judgement as an oracle for Type I hallucination occurrence.
When comparing \throne and CHAIR with human judgements,
we estimate using \throne improves the precision of judging Type I hallucinations
to 96\% versus 91\% when using CHAIR---this \textit{reduces the false discovery rate by more than 50\%}.
Note that we find most estimated errors in \throne arise from the particular class definitions in COCO,
\eg~the COCO class \texttts{tv} includes computer monitors, which the oracle judgement is aware of.

Second, we apply the same class (and image) sampling strategy as in POPE to \throne
and show this sampling overestimates $F^{0.5}_{\textsc{CLS}}$ by an average of $12.3$ points
compared to the complete use of classes in \throne~(\cref{tab:results_main}).

Finally, we vary the choice of $k$ \ie~the voting mechanism used to combine answers from multiple (LM, question) pairs.
We use the \emph{unanimous} voting mechanism ($k=9$) in \throne to minimize the false discovery rate
and find the valid alternatives of \emph{simple majority} ($k = 5$) or \emph{all-but-one} ($k = 8$) voting mechanisms
have strong correlations and rank correlations across all compute metrics, in \throne, of $>0.99$ and $>0.94$, respectively.

\cvprsection{Improved Baselines}\label{sec:imp_bl}
\vspace{-0.1cm}
% \prannay{Might need a some intro stuff here and work on linkage}
% In~\cref{ssec:results_ours},
% we show three successor LVLMs which achieves improved performance over its preceding version (MiniGPT4$\rightarrow$MiniGPT-v2, LLaMA-Adapter-v2$\rightarrow$v2.1, LLaVA-v1.3$\rightarrow$LLaVA-v1.5),
% with respect to Type II hallucinations on POPE (\cref{fig:results_pope}).
% However, these same models do not show any improvement in performance on \throne (\cref{tab:results_main}).
% This demonstrates addressing Type I hallucinations requires a different approach from existing methods that mitigate Type II hallucinations.
% Using LLaVA models~\cite{liu2023visual,liu2023improved},
Much needs to be done to study Type I hallucinations.
As a first step to their mitigation,
we demonstrate a baseline method to augment the visual instruction tuning data for LLaVA models~\cite{liu2023visual,liu2023improved},
yielding improvements on \throne while maintaining similar performance regarding Type II hallucinations on \textsc{POPE}.

\cvprsubsection{Visual Instruction Tuning Data Augmentation}
% To improve Type II hallucination performance,
Similar to chain-of-thought learning~\cite{wei2022chain},
during instruction tuning, we augment all visual instruction tuning samples constructed by LLaVA~\cite{liu2023visual}
by prepending the task of enumerating a list of objects (present and absent) and indicating approximate locations, if applicable.
% In particular, we prepend all visual instruction tuning data samples constructed by LLaVA~\cite{liu2023visual} with our object enumeration task.
Other than this, the LLaVA data and training pipeline remains unchanged.
To generate the new data for this object enumeration task,
we use the same COCO bounding box annotations used to generate the vision instruction tuning data in LLaVA.
The simple text-only format (no special tokens) we use is:

% \vspace{-0.2cm}
{\begingroup
\scriptsize
\begin{verbatim}
Instruction: <image> Give a list of objects and locations
             in the image.
Response:    {class_name_1} [{location_1}/absent]
             ...
             {class_name_N} [{location_N}/absent]
\end{verbatim}
\endgroup}
% \vspace{-0.3cm}
where \texttts{location\_i} is a plain text indicator representing the location of the center point of
the relevant object in the image on a $3\times3$ grid (\eg~\texttts{ bottom left}).
To provide negatives in the training data, if \texttts{class\_name\_i} is not present, we use the plain text indicator \texttts{absent}.
Prior work~\cite{zhou2023analyzing} shows that classes that frequently co-occur in the training data are the most common hallucinations,
therefore we bias our negative sampling towards class pairs that frequently co-occur using a correlation matrix.
We detail and ablate this choice in the Supplementary Material.
% \prannay{zhizhong to read}
% \zhizhong{This section really needs to be fleshed out... Or add a lot of pointers to SM. What is done in inference? What's the result? Do we ablate these choices?}

\cvprsubsection{Improved Baseline Results}\label{ssec:results_imp}
\begin{table}[t]
\centering
\scriptsize
\begin{adjustbox}{width=\linewidth}
\setlength\tabcolsep{2pt}
\begin{tabular}{l|c|ccc|ccc|ccc}
\toprule
                             & Object                              & \multicolumn{3}{c|}{\throne}                                                                                                                & \multicolumn{3}{c|}{POPE}                                                                                                                    & \multicolumn{3}{c}{POPE-C}                                                                                                                  \\
\multirow{-2}{*}{Model}      & Enumeration Data                    & $P_\textsc{CLS}$                     & $R_\textsc{CLS}$                     & \cellcolor[HTML]{EFEFEF}$F^{0.5}_{\textsc{CLS}}$             & $P$                                  & $R$                                  & \cellcolor[HTML]{EFEFEF}$F^{1}$                                & $P$                                  & $R$                                  & \cellcolor[HTML]{EFEFEF}$F^{1}$                              \\ \midrule
                             & \xmark                              & 79.9                                 & 65.3                                 & \cellcolor[HTML]{EFEFEF}76.5                                 & 58.0                                 & {\color[HTML]{0432FF} \textbf{98.4}} & \cellcolor[HTML]{EFEFEF}73.0                                   & 7.7                                  & {\color[HTML]{0432FF} \textbf{99.2}} & \cellcolor[HTML]{EFEFEF}14.3                                 \\
                             & COCO                              & 83.2                                 & {\color[HTML]{0432FF} \textbf{68.8}} & \cellcolor[HTML]{EFEFEF}79.9                                 & 73.2                                 & 88.2                                 & \cellcolor[HTML]{EFEFEF}80.0                                   & 9.8                                  & 69.4                                 & \cellcolor[HTML]{EFEFEF}17.2                                 \\
\multirow{-3}{*}{LLaVA-v1.3} & COCO + VG                         & {\color[HTML]{0432FF} \textbf{86.2}} & 67.0                                 & \cellcolor[HTML]{EFEFEF}{\color[HTML]{0432FF} \textbf{81.5}} & {\color[HTML]{0432FF} \textbf{83.0}} & 82.5                                 & \cellcolor[HTML]{EFEFEF}{\color[HTML]{0432FF} \textbf{82.8}}   & {\color[HTML]{0432FF} \textbf{13.8}} & 50.4                                 & \cellcolor[HTML]{EFEFEF}{\color[HTML]{0432FF} \textbf{21.7}} \\ \midrule
                             & \xmark                              & 69.9                                 & 56.4                                 & \cellcolor[HTML]{EFEFEF}66.8                                 & 81.9                                 & 90.8                                 & \cellcolor[HTML]{EFEFEF}86.1                                   & 58.7                                 & 85.7                                 & \cellcolor[HTML]{EFEFEF}69.7                                 \\
                             & COCO                              & {\color[HTML]{0432FF} \textbf{87.2}} & 76.6                                 & \cellcolor[HTML]{EFEFEF}{\color[HTML]{0432FF} \textbf{84.9}} & 88.6                                 & {\color[HTML]{0432FF} \textbf{85.3}} & \cellcolor[HTML]{EFEFEF}{\color[HTML]{0432FF} \textbf{87.0}}   & 58.9                                 & {\color[HTML]{0432FF} \textbf{87.5}} & \cellcolor[HTML]{EFEFEF}70.4                                 \\
\multirow{-3}{*}{LLaVA-v1.5} & COCO + VG                         & 86.1                                 & {\color[HTML]{0432FF} \textbf{77.0}} & \cellcolor[HTML]{EFEFEF}84.1                                 & {\color[HTML]{0432FF} \textbf{89.8}} & 83.7                                 & \cellcolor[HTML]{EFEFEF}86.7                                   & {\color[HTML]{0432FF} \textbf{64.5}} & 86.1                                 & \cellcolor[HTML]{EFEFEF}{\color[HTML]{0432FF} \textbf{73.7}} \\ \bottomrule
\end{tabular}
\end{adjustbox}
\caption{\textbf{Improved Baseline via Object Enumeration:} Adding our object enumeration task to LLaVA training and inference leads to large improvements on \throne particularly
in terms of classwise precision, $P_\textsc{CLS}$, over standard LLaVA models, demonstrating a reduction in Type I hallucinations,
as well as small reductions in Type II hallucinations judged by POPE and POPE-C.
}\label{tab:results_improve}
\vspace{-0.5cm}
\end{table}

\cref{tab:results_improve} shows the result of evaluating our improved baseline method on \throne, POPE, and POPE-C.
During inference, we approximate our training data augmentation by first prompting the LVLM to perform the object enumeration tasks \emph{and then}
generating a response to the prompt from the relevant benchmark.
We additionally show results of utilizing VisualGenome bounding box annotations~\cite{krishna2017visual} on the COCO images where available. 
On \throne, we observe a large increase in classwise precision and therefore $F^{0.5}_{\textsc{CLS}}$, particularly for LLaVA-v1.5,
demonstrating the ability of our method to reduce Type I hallucinations.
Moreover, on POPE and POPE-C, using our object enumeration yields small improvements in precision,
indicating reduced Type II hallucinations as well.
In the Supplementary Material, we ablate the sampling of negatives during object enumeration training
and the effect of removing the object enumeration task during inference.
\cvprsection{Conclusion}\label{sec:conc}
We establish a novel benchmark, \throne, for evaluating hallucinations
generated by LVLMs in free-form image descriptions \ie~\emph{Type I hallucinations}.
Our benchmark utilizes multiple LMs and prompt formats with a simple voting mechanism to yield
an accurate evaluation of Type I hallucinations in LVLM responses.
We ensure that \throne is broadly accessible by utilizing open-source LMs capable of running on a single commercial GPU.
Using \throne, we benchmark 11 publicly available LVLMs on two datasets, COCO and Objects365,
and demonstrate that limited progress has been made in addressing Type I hallucinations.
% Using \throne on COCO and Objects365, we show that much additional work is needed to address Type I hallucinations.
Moreover, we show how the established benchmark, POPE, underestimates \emph{Type II hallucinations}, which occur in response to specific questions
\eg~yes-no questions. We present results for a completed version (POPE-C) to enable a comparison of Type I hallucinations through \throne
and Type II hallucinations using POPE-C.
Finally, we propose a simple data augmentation for LVLM training that can result in a large reduction in Type I hallucinations whilst
maintaining or improving Type II hallucination performance. % \yoni{double check if you mean type I and type II in the last sentence}

Limitations and Ethical Considerations are discussed in the Supplementary Material.% \zhizhong{This is duplicating the introduction's contributions.}

\clearpage

{
    \small
    \bibliographystyle{ieeenat_fullname}
    \bibliography{longstrings,citations}
}


% \newpage
\clearpage
\onecolumn
\appendix

\newcommand*{\smpath}{supplementary_material/sm_sec}%
\newcommand*{\smassp}{supplementary_material/sm_assets}%

% \input{sm_sec/01-model_details.tex}

% \clearpage

\cvprsection{Voting Mechanism Ablation}\label{smsec:voting_mech_ablation}
\begin{table*}[h]
    \centering
    \begin{tabular}{l|c|cccc|cccc|c}
    \toprule
    Model                               & $k$ & $P_{\text{ALL}}$ & $R_{\text{ALL}}$ & $F_{\text{ALL}}^{1}$ & $F_{\text{ALL}}^{0.5}$ & $P_{\text{CLS}}$ & $R_{\text{CLS}}$ & $F_{\text{CLS}}^{1}$ & $F_{\text{CLS}}^{0.5}$ & Ignore \% \\ \midrule
    \multirow{3}{*}{Adapter-v2}   & 9   & 63.6             & 73.3             & 68.1                 & 65.3                   & 68.2             & 70.6             & 69.4                 & 68.7                   & 2.4       \\
                                        & 5   & 61.8             & 75.0             & 67.7                 & 64.0                   & 65.7             & 72.0             & 68.7                 & 66.9                   & 0.0       \\
                                        & 8   & 63.4             & 73.8             & 68.2                 & 65.2                   & 68.0             & 70.8             & 69.4                 & 68.5                   & 1.4       \\ \midrule
    \multirow{3}{*}{Adapter-v2.1} & 9   & 63.8             & 73.7             & 68.4                 & 65.5                   & 67.4             & 71.2             & 69.3                 & 68.1                   & 2.4       \\
                                        & 5   & 61.7             & 75.3             & 67.8                 & 64.0                   & 64.7             & 72.5             & 68.4                 & 66.1                   & 0.0       \\
                                        & 8   & 63.6             & 74.1             & 68.5                 & 65.5                   & 67.2             & 71.5             & 69.3                 & 68.0                   & 1.5       \\ \midrule
    \multirow{3}{*}{InstructBLIP}       & 9   & 70.8             & 74.3             & 72.5                 & 71.5                   & 77.2             & 71.9             & 74.5                 & 76.1                   & 2.5       \\
                                        & 5   & 68.2             & 77.2             & 72.4                 & 69.8                   & 73.2             & 74.3             & 73.7                 & 73.4                   & 0.0       \\
                                        & 8   & 70.6             & 75.1             & 72.8                 & 71.4                   & 76.8             & 72.3             & 74.5                 & 75.9                   & 1.5       \\ \midrule
    \multirow{3}{*}{Otter-Image}        & 9   & 33.0             & 31.2             & 32.1                 & 32.7                   & 25.2             & 16.9             & 20.2                 & 22.9                   & 8.5       \\
                                        & 5   & 25.6             & 34.7             & 29.5                 & 27.1                   & 16.4             & 20.1             & 18.0                 & 17.0                   & 0.0       \\
                                        & 8   & 32.4             & 31.8             & 32.1                 & 32.3                   & 23.9             & 17.2             & 20.0                 & 22.2                   & 4.8       \\ \midrule
    \multirow{3}{*}{MiniGPT-4}          & 9   & 81.7             & 59.8             & 69.0                 & 76.1                   & 79.9             & 61.8             & 69.7                 & 75.5                   & 2.9       \\
                                        & 5   & 74.8             & 64.9             & 69.5                 & 72.6                   & 73.0             & 65.4             & 69.0                 & 71.3                   & 0.0       \\
                                        & 8   & 80.8             & 61.1             & 69.6                 & 75.9                   & 79.1             & 62.4             & 69.8                 & 75.1                   & 1.8       \\ \midrule
    \multirow{3}{*}{MiniGPT-v2}         & 9   & 79.0             & 66.6             & 72.3                 & 76.2                   & 77.6             & 67.0             & 71.9                 & 75.2                   & 2.8       \\
                                        & 5   & 73.6             & 71.4             & 72.5                 & 73.1                   & 72.1             & 70.5             & 71.3                 & 71.7                   & 0.0       \\
                                        & 8   & 78.4             & 67.8             & 72.7                 & 76.0                   & 76.9             & 67.7             & 72.0                 & 74.8                   & 1.8       \\ \midrule
    \multirow{3}{*}{LLaVA-Mistral}       & 9   & 86.8             & 71.8             & 78.3                 & 83.6                   & 84.4             & 64.2             & 70.8                 & 77.5                   & 2.7       \\
                                        & 5   & 82.8             & 75.9             & 78.3                 & 81.2                   & 78.5             & 68.3             & 71.2                 & 77.4                   & 0.0       \\
                                        & 8   & 86.3             & 73.1             & 78.5                 & 83.3                   & 83.7             & 65.0             & 71.2                 & 77.4                   & 1.6       \\ \midrule
    \multirow{3}{*}{mPLUG-Owl}          & 9   & 55.5             & 71.9             & 62.6                 & 58.1                   & 66.3             & 68.3             & 67.3                 & 66.7                   & 2.4       \\
                                        & 5   & 54.3             & 73.9             & 62.6                 & 57.3                   & 63.7             & 69.9             & 66.6                 & 64.8                   & 0.0       \\
                                        & 8   & 55.5             & 72.6             & 62.9                 & 58.2                   & 66.2             & 68.6             & 67.4                 & 66.7                   & 1.4       \\ \midrule
    \multirow{3}{*}{LRV-Instruction-v2} & 9   & 82.0             & 56.7             & 67.0                 & 75.3                   & 78.4             & 58.8             & 67.2                 & 73.5                   & 3.6       \\
                                        & 5   & 77.5             & 60.8             & 68.1                 & 73.5                   & 74.6             & 61.9             & 67.7                 & 71.7                   & 0.0       \\
                                        & 8   & 81.7             & 57.9             & 67.8                 & 75.5                   & 78.0             & 59.4             & 67.4                 & 73.4                   & 2.0       \\ \midrule
    \multirow{3}{*}{LLaVA-v1.3}         & 9   & 80.5             & 65.2             & 72.1                 & 76.9                   & 79.9             & 65.3             & 71.9                 & 76.5                   & 2.4       \\
                                        & 5   & 76.4             & 68.7             & 72.4                 & 74.7                   & 75.6             & 68.0             & 71.6                 & 73.9                   & 0.0       \\
                                        & 8   & 80.0             & 66.3             & 72.5                 & 76.9                   & 79.4             & 65.8             & 72.0                 & 76.3                   & 1.4       \\ \midrule
    % \multirow{3}{*}{LLaVA-Mistral}         & 9   & 68.1             & 61.0             & 64.4                 & 66.6                   & 69.9             & 56.4             & 62.5                 & 66.8                   & 2.7       \\
    %                                     & 5   & 65.5             & 62.6             & 64.0                 & 64.9                   & 67.1             & 57.8             & 62.1                 & 65.0                   & 0.0       \\
    %                                     & 8   & 67.8             & 61.5             & 64.5                 & 66.4                   & 69.6             & 56.7             & 62.5                 & 66.6                   & 1.3       \\ \bottomrule
    \\ \bottomrule
    \end{tabular}
    \caption{\textbf{Comparison of Voting Mechanisms in \textsc{THRONE}.}
    We compare three different voting mechanisms: \emph{unanimous}, $k=9$; \emph{simple majority}, $k=5$; and \emph{jury majority}, $k=8$.
    Moreover in each case we report the number of ignore labels as a result of each voting mechanism. The number of labels is $400,000$ for a given LVLM.
    We find that the number of ignore labels is low is almost all cases and metrics are strongly correlated ($>0.99$).
    In \textsc{THRONE}, we use a \emph{unanimous} voting mechanism ($k=9$) to minimize the likelihood of hallucination judgement errors.
    }\label{smtab:voting_ablation}
\end{table*}

\clearpage

\cvprsection{\throne vs. POPE Sampling}\label{smsec:throne_vs_samp}
\begin{table}[h!]
    \centering
    \begin{tabular}{l|c|cccc|cccc}
    \toprule
    Model                                & Sampling    & $P_{\text{ALL}}$               & $R_{\text{ALL}}$              & $F_{\text{ALL}}^{1}$           & $F_{\text{ALL}}^{0.5}$         & $P_{\text{CLS}}$               & $R_{\text{CLS}}$              & $F_{\text{CLS}}^{1}$           & $F_{\text{CLS}}^{0.5}$         \\ \midrule
                                         & THRONE      & 63.6                           & 73.3                          & 68.1                           & 65.3                           & 68.2                           & 70.6                          & 68.3                           & 68.0                           \\
                                         & POPE        & 77.3                           & 73.2                          & 75.2                           & 76.5                           & 84.6                           & 70.0                          & 76.9                           & 81.2                           \\
    \multirow{-3}{*}{Adapter-v2}   & $\Delta$    & {\color[HTML]{FF0000} (-13.7)} & 0.2                           & {\color[HTML]{FF0000} (-7.1)}  & {\color[HTML]{FF0000} (-11.1)} & {\color[HTML]{FF0000} (-16.3)} & 0.5                           & {\color[HTML]{FF0000} (-8.6)}  & {\color[HTML]{FF0000} (-13.2)} \\ \midrule
                                         & THRONE      & 63.8                           & 73.7                          & 68.4                           & 65.5                           & 67.4                           & 71.2                          & 68.1                           & 67.5                           \\
                                         & POPE        & 79.2                           & 73.2                          & 76.1                           & 77.9                           & 85.4                           & 71.8                          & 76.3                           & 80.6                           \\
    \multirow{-3}{*}{Adapter-v2.1} & $\Delta$    & {\color[HTML]{FF0000} (-15.4)} & 0.5                           & {\color[HTML]{FF0000} (-7.7)}  & {\color[HTML]{FF0000} (-12.4)} & {\color[HTML]{FF0000} (-18.0)} & {\color[HTML]{FF0000} (-0.6)} & {\color[HTML]{FF0000} (-8.2)}  & {\color[HTML]{FF0000} (-13.2)} \\ \midrule
                                         & THRONE      & 70.8                           & 74.3                          & 72.5                           & 71.5                           & 77.2                           & 71.9                          & 73.1                           & 75.2                           \\
                                         & POPE        & 82.2                           & 74.4                          & 78.1                           & 80.5                           & 85.0                           & 70.0                          & 77.9                           & 82.6                           \\
    \multirow{-3}{*}{InstructBLIP}       & $\Delta$    & {\color[HTML]{FF0000} (-11.4)} & 0.0                           & {\color[HTML]{FF0000} (-5.5)}  & {\color[HTML]{FF0000} (-9.0)}  & {\color[HTML]{FF0000} (-7.8)}  & 1.9                           & {\color[HTML]{FF0000} (-4.8)}  & {\color[HTML]{FF0000} (-7.4)}  \\ \midrule
                                         & THRONE      & 33.0                           & 31.2                          & 32.1                           & 32.7                           & 25.2                           & 16.9                          & 18.7                           & 21.5                           \\
                                         & POPE        & 66.5                           & 34.9                          & 45.7                           & 56.3                           & 70.7                           & 17.3                          & 35.6                           & 49.2                           \\
    \multirow{-3}{*}{Otter-Image}        & $\Delta$    & {\color[HTML]{FF0000} (-33.5)} & {\color[HTML]{FF0000} (-3.7)} & {\color[HTML]{FF0000} (-13.7)} & {\color[HTML]{FF0000} (-23.6)} & {\color[HTML]{FF0000} (-45.5)} & {\color[HTML]{FF0000} (-0.4)} & {\color[HTML]{FF0000} (-17.0)} & {\color[HTML]{FF0000} (-27.7)} \\ \midrule
                                         & THRONE      & 81.7                           & 59.8                          & 69.0                           & 76.1                           & 79.9                           & 61.8                          & 67.6                           & 73.6                           \\
                                         & POPE        & 89.1                           & 58.7                          & 70.8                           & 80.7                           & 88.2                           & 60.0                          & 70.5                           & 78.9                           \\
    \multirow{-3}{*}{MiniGPT-4}          & $\Delta$    & {\color[HTML]{FF0000} (-7.5)}  & 1.1                           & {\color[HTML]{FF0000} (-1.7)}  & {\color[HTML]{FF0000} (-4.7)}  & {\color[HTML]{FF0000} (-8.2)}  & 1.8                           & {\color[HTML]{FF0000} (-2.9)}  & {\color[HTML]{FF0000} (-5.4)}  \\ \midrule
                                         & THRONE      & 79.0                           & 66.6                          & 72.3                           & 76.2                           & 77.6                           & 67.0                          & 70.4                           & 74.0                           \\
                                         & POPE        & 88.3                           & 65.8                          & 75.4                           & 82.7                           & 89.8                           & 68.5                          & 76.3                           & 82.6                           \\
    \multirow{-3}{*}{MiniGPT-v2}         & $\Delta$    & {\color[HTML]{FF0000} (-9.3)}  & 0.8                           & {\color[HTML]{FF0000} (-3.1)}  & {\color[HTML]{FF0000} (-6.5)}  & {\color[HTML]{FF0000} (-12.2)} & {\color[HTML]{FF0000} (-1.5)} & {\color[HTML]{FF0000} (-5.9)}  & {\color[HTML]{FF0000} (-8.6)}  \\ \midrule
                                         & THRONE      & 74.7                           & 77.2                          & 75.9                           & 75.2                           & 78.0                           & 76.1                          & 76.3                           & 77.1                           \\
                                         & POPE        & 83.8                           & 76.6                          & 80.0                           & 82.2                           & 87.8                           & 74.6                          & 80.0                           & 84.4                           \\
    \multirow{-3}{*}{LLaVA-Mistral}       & $\Delta$    & {\color[HTML]{FF0000} (-9.1)}  & 0.7                           & {\color[HTML]{FF0000} (-4.1)}  & {\color[HTML]{FF0000} (-7.1)}  & {\color[HTML]{FF0000} (-9.8)}  & 1.5                           & {\color[HTML]{FF0000} (-3.7)}  & {\color[HTML]{FF0000} (-7.3)}  \\ \midrule
                                         & THRONE      & 55.5                           & 71.9                          & 62.6                           & 58.1                           & 66.3                           & 68.3                          & 65.2                           & 65.3                           \\
                                         & POPE        & 72.9                           & 71.2                          & 72.0                           & 72.6                           & 82.0                           & 64.5                          & 74.1                           & 78.6                           \\
    \multirow{-3}{*}{mPLUG-Owl}          & $\Delta$    & {\color[HTML]{FF0000} (-17.4)} & 0.7                           & {\color[HTML]{FF0000} (-9.4)}  & {\color[HTML]{FF0000} (-14.4)} & {\color[HTML]{FF0000} (-15.6)} & 3.7                           & {\color[HTML]{FF0000} (-8.9)}  & {\color[HTML]{FF0000} (-13.3)} \\ \midrule
                                         & THRONE      & 82.0                           & 56.7                          & 67.0                           & 75.3                           & 78.4                           & 58.8                          & 65.0                           & 71.5                           \\
                                         & POPE        & 88.6                           & 54.8                          & 67.7                           & 78.9                           & 85.0                           & 56.2                          & 68.8                           & 77.4                           \\
    \multirow{-3}{*}{LRV-Instruction-v2} & $\Delta$    & {\color[HTML]{FF0000} (-6.6)}  & 1.9                           & {\color[HTML]{FF0000} (-0.7)}  & {\color[HTML]{FF0000} (-3.6)}  & {\color[HTML]{FF0000} (-6.6)}  & 2.6                           & {\color[HTML]{FF0000} (-3.7)}  & {\color[HTML]{FF0000} (-5.9)}  \\ \midrule
                                         & THRONE      & 80.5                           & 65.2                          & 72.1                           & 76.9                           & 79.9                           & 65.3                          & 70.4                           & 75.2                           \\
                                         & POPE        & 85.5                           & 61.6                          & 71.6                           & 79.3                           & 87.9                           & 61.7                          & 72.6                           & 80.8                           \\
    \multirow{-3}{*}{LLaVA-v1.3}         & $\Delta$    & {\color[HTML]{FF0000} (-4.9)}  & 3.6                           & 0.5                            & {\color[HTML]{FF0000} (-2.4)}  & {\color[HTML]{FF0000} (-8.0)}  & 3.6                           & {\color[HTML]{FF0000} (-2.2)}  & {\color[HTML]{FF0000} (-5.5)}  \\ \midrule
                                         & THRONE      & 68.1                           & 61.0                          & 64.4                           & 66.6                           & 69.9                           & 56.4                          & 62.2                           & 66.8                           \\
                                         & POPE        & 81.5                           & 64.4                          & 72.0                           & 77.4                           & 86.2                           & 59.2                          & 70.4                           & 78.8                           \\
    \multirow{-3}{*}{LLaVA-v1.5}         & $\Delta$    & {\color[HTML]{FF0000} (-13.4)} & {\color[HTML]{FF0000} (-3.4)} & {\color[HTML]{FF0000} (-7.6)}  & {\color[HTML]{FF0000} (-10.9)} & {\color[HTML]{FF0000} (-16.2)} & {\color[HTML]{FF0000} (-2.7)} & {\color[HTML]{FF0000} (-8.2)}  & {\color[HTML]{FF0000} (-12.0)} \\ \bottomrule
    \end{tabular}
    \caption{\textbf{Balanced Sampling (POPE) vs. Exhaustive Sampling (\textsc{THRONE}):}
    Applying POPE sampling to \textsc{THRONE} leads to an underestimation of the prevalence
    of Type I hallucinations regardless of LVLM.\@
    }\label{smtab:pope_thro_sampling}
\end{table}

In the main paper,
we demonstrated how the sampling method used in POPE~\cite{li2023evaluating} leads to an underestimation of Type II hallucinations
and outline a complete version of POPE (POPE-C), which shows the true extent of Type II hallucinations in LVLMs.
In this section, we perform the opposite---we apply POPE type sampling to \throne and compare the results to the complete sampling
method used in \throne as outlined in the main paper.
\cref{smtab:pope_thro_sampling} shows the results of applying POPE style sampling to \throne,
once again, applying POPE style sampling leads to a large underestimation of Type I hallucinations.
POPE style sampling applied to \throne leads to a mean underestimation of $F_{\text{CLS}}^{0.5}$ by $10.9$ points
compared to complete sampling, which is the default in \throne.

% \clearpage

\cvprsection{CHAIR Overview}\label{smsec:chair}
% \lipsum[1]
We present a detailed description of the CHAIR evaluation presented in \cite{rohrbach2018object} below. The method of CHAIR, like \throne, attempts to capture the extent of hallucinations in free-form generated text pertaining an image, however, focuses on more traditional image captioners. Similarly to \throne, CHAIR does not use concept-focused prompts (\ie instead is focused on ``Type-I'') and is intended to be used only in captioning tasks. CHAIR defines a manual pipeline on-top of the annotated MSCOCO image dataset to produce a list of ground truth objects present in the scene. 

Given a set of ground truth objects in an image and a model-generated image caption, CHAIR extracts the objects present in the scene via a traditional hard-rule extraction method and then attempts to map each of the predicted objects into one of the 80 class categories of MSCOCO. The mapping of the extracted objects from the caption into the class set of MSCOCO uses a pre-defined synonym dictionary for each object category. Once the objects of the predicted caption are extracted and mapped to one of the MSCOCO categories, CHAIR evaluates for ``false-positive'' predictions (\ie hallucinations). In particular the authors introduce two variants of CHAIR, the first variant $\text{CHAIR}_i$ quantifies the extent of hallucinated instances as,
\begin{equation*}
\text{CHAIR}_i = \frac{|\{ \text{hall. object} \}|}{|\{ \text{pred. object} \}|}.
\end{equation*}
In this setting, hallucinations would be objects extracted from the model-generated captions that after being mapped, are still not present in the ground-truth object list for the corresponding instance. We note that $\text{CHAIR}_i$ can be viewed as measuring the ``false discovery rate'' (FDR) that is $1-P$ where $P$ is the precision. Using Precision as the main metric, however is limiting as it does not take into account the ``False Negative Rate'' (FNR) which is $1-R$ where $R$ is the recall. This implies that by lacking recall measurements, $\text{CHAIR}_i$ may assign high scores to short and incomplete captions which are not comprehensive in detailing the image. This is in stark contrast with the new generation of  LVLM models powered by LLMs which are designed to be more exhaustive and detailed and makes the use of $\text{CHAIR}_i$ problematic when evaluating with LVLMs. We note that $\text{CHAIR}_i$ is the main metric used when people report ``$\text{CHAIR}$ scores'', and typically reported numbers correspond to the mean $\text{CHAIR}_i$ score across the MSCOCO validation set of images.

The second variation of $\text{CHAIR}$ is the $\text{CHAIR}_s$ which simply measure the number of sentences (predictions) that include at least 1 hallucination as compared with all sentences considered, 
\begin{equation}
    \text{CHAIR}_s = \frac{|\{ \text{sentences w/ hall. object} \}|}{|\{ \text{all sentences} \}|}.
\end{equation}
Note that $\text{CHAIR}_s$ does not measure the extent of hallucination within a sentence, just the existence of at least one hallucination. This is problematic as it does not capture the extent of hallucination in the sentence especially for long-form text and does not elucidate if a caption contains many or a single hallucination.

\paragraph{Producing ground truth in CHAIR} To create a list of ground truth objects from MSCOCO annotations, the authors of CHAIR harness two annotation types to produce the most exhaustive list of ground truth objects. First the authors directly use all of the instance segmentation labels for each image, which they aggregate into a unique list of objects existing in the image. Next the authors use the 5 human-labelled captions of each image in MSCOCO and use the same extraction and mapping pipeline applied to the predictions to produce an additional set of objects that are present in the captions. Both objects lists are combined and the authors note that captioning ground truth and instance segmentation ground truth objects are often complementary as they follow different styles. Therefore combining objects from both types of object lists is beneficial for the most exhaustive final ground truth list.

% \paragraph{Ethical considerations}

% method of measuring hallucinations in LVLMs and as such is helping the community in progressing forward in developing LVLMs that can be rellied on critical  settings. 

% ethical considerations
% controlling the use of LVLMs to be used in reliable applications
% As with many computer vision, there is a risk of surveillence, however this work does not focus on this at all...
% improving model evaluation helps practioners compare existing models and pick up the best one...

% limitations
% This paper does not claim to solve the problem of hallucination
% hallucination goes past the existnce of objects in a scene, despite that we are catching abstract descriptions see fig. x still not captuing all types of hallucinations
% this paper does not measure bias of LVLMs of particular types
% presented method focuses on type I, we believe both type i and ii are important and as we observe are contradictory

% \clearpage

\cvprsection{Qualitative Evaluation of \throne}\label{smsec:qual_eval_throne}

\cvprsubsection{Evaluation Method}
\begin{table}[t]
    \centering
    \begin{adjustbox}{width=\linewidth}
    \begin{tabular}{@{}l|c|c|ccc|c@{}}
    \toprule
    Method                & CHAIR == \textsc{THRONE}? & \#Responses (\%)  & \#Responses Evaluated & \#Judgements & \#Judgement Errors & Error Rate \\ \midrule
    CHAIR/THRONE          & \checkmark               & 38350 (69.7\%)     & 55                    & 157          & 4                  & 2.5\%      \\
    \textsc{THRONE}       & \xmark                   & 16650 (30.3\%)     & 110                   & 376          & 30                 & 8.0\%      \\
    CHAIR                 & \xmark                   & 16650 (30.3\%)     & 110                   & 477          & 111                & 23.3\%     \\ \bottomrule
    \end{tabular}
\end{adjustbox}
\caption{\textbf{Summary of Qualitative Evaluation:} Our qualitative results show that for responses in which \textsc{THRONE} and CHAIR
differ, there is a large difference in the error rate. When \textsc{THRONE} and CHAIR agree, the error rate is small.
}\label{smtab:qual_eval}
\end{table}

\begin{table}[t]
    \centering
    \begin{tabular}{@{}c|c|c|c@{}}
        \toprule
        \#Responses Analyzed & \#False Positives Identified & \#Hallucinations & \#Misclassifications \\ \midrule
        90                   & 71                           & 69               & 2                    \\ \bottomrule
    \end{tabular}
    \caption{\textbf{Hallucinations Dominate False Positives:} Human evaluation establishes the vast majority ($\frac{69}{71}\approx 97\%$) of
    false positive object classes in LVLM responses are true hallucinations rather than plausible misclassifications of objects.
    }\label{smtab:qual_hal}
\end{table}
We include a self-contained file (\texttts{THRONE\_qual\_eval\_results.html}) in the Supplementary Material,
which shows the qualitative evaluation and comparison of \throne and CHAIR.\@
For each LVLM evaluated, we sample $10$ COCO images at random in which \throne and CHAIR \emph{disagree}
and $5$ COCO images in which \throne and CHAIR \emph{agree}.
Therefore, we qualitatively evaluate $165$ responses.
These results are summarized in~\cref{smtab:qual_eval}

By calculating error rates for each of these cases and noting the proportion of responses in which \throne and CHAIR \emph{disagree}
we can estimate the overall error rate of each method using a weighted sum.
\begin{align*}
    \textbf{Method Error Rate} = 
    & \ (\text{Agreement Proportion}) \times (\text{Error Rate in Agreement Case}) \\
    & + (\text{Disagreement Proportion}) \times (\text{Error Rate in Disagreement Case}) \\
    \textbf{CHAIR Error Rate} =
    & \ 0.697 \times 0.025 + 0.303 \times 0.233 = 0.088 = 8.8\% \\
    \textbf{\throne Error Rate} =
    & \ 0.697 \times 0.025 + 0.303 \times 0.080 = 0.043 = 4.3\%
\end{align*}

\cvprsubsection{Discussion}
We find that the plurality of errors made in \throne relate to a mismatch between the LM definition of a certain class and
the definiton in COCO.\@
The most clear example is in the \texttts{tv} COCO class.
In COCO, this class includes computer monitors,
whereas for an LM, the implication of the existence computer monitors in an LVLM response does not lead to a ``\texttts{yes}'' response
when asked \texttts{Is there a tv in this image?} or similar.
When doing an manual evaluation our human oracle \emph{is} aware of the particular COCO class definitions and answers accordingly.
Using a handcrafted rule for \texttts{tv} and other similar COCO classes, we would expect the error rate of \throne to reduce
significantly, but in \throne we deliberately avoid the use of handcrafted rules.

As mentioned in the main paper, the errors in CHAIR are more fundamental and result due to simple text matching of synonyms not being able
to discriminate between abstract concepts alluded to in a response and direct objects implied to exist in the image based on the response.

\cref{smtab:qual_hal} shows results for human analysis of false positives.
We analyze $90$ responses, the $15$ samples for mPLUG-Owl, MiniGPT-v2, MiniGPT-4, LLaVA-7b-v1.5, LLaVA-7b-v1.3 and InstructBLIP---the
final $90$ responses in the self-contained file: \texttts{THRONE\_qual\_eval\_results.html} in the Supplementary Material.

% \clearpage

\cvprsection{Improved Baseline Implementation Details}\label{smsec:imp_baseline}
In the main paper, we introduced a simple method to augment the LLaVA visual instruction
tuning data with an object enumeration task to reduce Type I and Type II hallucinations
when used to train LLaVA models.
The format used for object enumeration is:

\begin{verbatim}
Instruction: <image> Give a list of objects and locations
             in the image.
Response:    {class_name_1} [{location_1}/absent]
             ...
             {class_name_N} [{location_N}/absent]
\end{verbatim}
where \texttts{location\_i} represents the location of the bounding box center point on a $3\times 3$ grid.

We give additional details on the construction of the object enumeration task here.

\cvprsubsection{Object Enumeration Implementation Details}
The LLaVA visual instruction tuning data contains 157712 samples applied to 81479 images from the COCO training set
(some images correspond to multiple samples).
We ensure the absolute character length of our object enumerate task for a single sample is not exceedingly long---we
do not want the visual instruction tuning data to be pushed outside the context length of the LLaVA model.
This is done by limiting the number of instances \emph{per class} in a sample to 3.

For each \emph{sample} we construct an object enumeration task using bounding box data as follows:
first, sort bounding box annotations for a given image by box area in descending order;
second, loop over the sorted annotations adding the instance (\texttts{class\_name\_i, location\_i}) to the object enumeration task if
there are less than 3 instances of \texttts{class\_name} in the task;
third, sample 6 negative classes and append them to the object enumeration task using \texttts{absent} as the location string.

The sampling of negative classes is detailed next.

\cvprsubsection{Negative Sampling Implementation Details}
To sample negatives in the object enumeration task we first build a co-occurence matrix from the bounding box annotations.
The pseudocode for building this matrix is as follows:
\begin{verbatim}
from pycocotools.coco import COCO
import numpy as np
train_dset = COCO(instances_path)
num_cats = len(train_dset.getCatIds())
co_occur = np.array((num_cats, num_cats))
cat_id2cont_id = {x: i for i, x in sorted(enumerate(train_dset.getCatIds()))}
for iid in train_dset.getImgIds():
    anns = train_dset.loadAnns(train_dset.getAnnIds(imgIds=iid))
    pres_cats = [coco_cid2cont_cid[x['category_id']] for x in anns]
    pres_cats = np.unique(pres_cats)
    for r in pres_cats:
        for c in pres_cats:
            if r != c:
                co_occur[r, c] += 1
\end{verbatim}

After building this co-occurence matrix negative classes are sampled in
a manner which is aware of the classes present in a given image.
The pseudocode is as follows (using some variables from the above pseudocode):
\begin{verbatim}
present_cat_ids: List[int]  # list of category ids present in the image
present_cont_ids = [cat_id2cont_id[x] for x in present_cat_ids

# combine co-occurence across present categories
# ensuring present categories can not be sampled
present_co_occur = co_occur[present_cont_ids].copy()
present_co_occur[:, present_cont_ids] = 0
present_co_occur = present_co_occur / present_co_occur.sum(axis=1, keepdims=True)
present_co_occur = present_co_occur.sum(axis=0)
present_co_occur = present_co_occur / present_co_occur.sum()
# sharpen distribution
present_co_occur = present_co_occur ** 10
present_co_occur = present_co_occur / present_co_occur.sum()


rng = np.random.RandomState(iid)
neg_ids = rng.choice(
    sorted(train_dset.getCatIds()),
    size=6,
    p=present_co_occur,
    replace=False
)
\end{verbatim}

This method of sampling yields negative classes which commonly co-occur with positive classes in a given image.
Therefore, the object enumeration task trains the LVLM to distinguish individual objects and classes rather
than relying on global context.

\cvprsubsection{Object Enumeration Data Details}
In Table 3 of the main paper, we present results on \throne, POPE and POPE-C when training with our
object enumeration task using COCO or COCO and VisualGenome as object enumeration data.
Approximately 33000 of the 81479 COCO images in the LLaVA visual instruction tuning data are contained in the VisualGenome
dataset.
When using COCO \emph{and} VisualGenome data, we construct the object enumeration task for an image from VisualGenome
data when possible and COCO otherwise---we do not combine COCO and VisualGenome annotations for any image.

\cvprsubsection{Inference Details}
In Table 3 of the main paper, we present results on \throne, POPE and POPE-C when training with our
object enumeration task \emph{and} performing the object enumeration task at inference.
In the next section we show the effect of not performing object enumeration during inference on \throne and POPE,
instead directly addressing the relevant task.

\cvprsubsection{Ablation Results on Improved Baseline}
\begin{table*}[h!]
    \centering
    \scriptsize
    \begin{tabular}{c|c|c|c|ccc|ccc}
        \toprule
                                & Obj. Enum. & Obj. Enum. & Obj. Enum. & \multicolumn{3}{c|}{THRONE}                                                            & \multicolumn{3}{c}{POPE}                      \\
        \multirow{-2}{*}{Model} & Data       & Negatives  & Inference  & $P_\textsc{CLS}$ & $R_\textsc{CLS}$ & \cellcolor[HTML]{EFEFEF}$F^{0.5}_{\textsc{CLS}}$ & $P$  & $R$  & \cellcolor[HTML]{EFEFEF}$F^{1}$ \\ \midrule
                                & \xmark     & N/A        & N/A        & 79.9             & 65.3             & \cellcolor[HTML]{EFEFEF}76.5                     & 58.0 & 98.4 & \cellcolor[HTML]{EFEFEF}73.0    \\
                                & COCO       & \xmark     & \checkmark & 82.4             & 69.2             & \cellcolor[HTML]{EFEFEF}79.4                     & 64.8 & 95.2 & \cellcolor[HTML]{EFEFEF}77.1    \\
        LLaVA-v1.3              & COCO + VG  & \xmark     & \checkmark & 85.8             & 60.4             & \cellcolor[HTML]{EFEFEF}79.1                     & 66.0 & 95.3 & \cellcolor[HTML]{EFEFEF}78.0    \\
                                & COCO       & \checkmark & \checkmark & 83.2             & 68.8             & \cellcolor[HTML]{EFEFEF}79.9                     & 73.2 & 88.2 & \cellcolor[HTML]{EFEFEF}80.0    \\
                                & COCO + VG  & \checkmark & \checkmark & 86.2             & 67.0             & \cellcolor[HTML]{EFEFEF}81.5                     & 83.0 & 82.5 & \cellcolor[HTML]{EFEFEF}82.8    \\ \midrule
                                & \xmark     & N/A        & N/A        & 69.9             & 56.4             & \cellcolor[HTML]{EFEFEF}66.8                     & 81.9 & 90.8 & \cellcolor[HTML]{EFEFEF}86.1    \\
        LLaVA-v1.5              & COCO + VG  & \checkmark & \xmark     & 79.3             & 76.1             & \cellcolor[HTML]{EFEFEF}78.6                     & 83.2 & 86.4 & \cellcolor[HTML]{EFEFEF}84.8    \\
                                & COCO + VG  & \checkmark & \checkmark & 86.1             & 77.0             & \cellcolor[HTML]{EFEFEF}84.1                     & 89.8 & 83.7 & \cellcolor[HTML]{EFEFEF}86.7    \\ \bottomrule
        \end{tabular}
    \caption{\textbf{Effect of Negatives and Inference:} Including negatives in our object enumeration task improves performance on \textsc{THRONE} and POPE in terms of precision
    and F-score.
    Performing the object enumeration task at inference time improves performance on \textsc{THRONE} and POPE, but hampers inference time as the
    object enumeration task can generate long sequences.
    }\label{smtab:results_improve}
    \vspace{-0.5cm}
    \end{table*}

% \clearpage

\cvprsection{Limitations}\label{smsec:limitations}
% \lipsum[1]
% \paragraph{Limitations}
In this paper we present \throne which is a step towards measuring and mitigating hallucinations in LVLMs, nonetheless, our work has few key limitations which we list below.
\begin{enumerate}
    \item \throne is concerned with \emph{only} measuring hallucinations in LVLM predictions in the form of a \emph{false existence of an object in a closed set of classes}. As observed in LLMs, hallucinations are much more multifaceted and include not just objects outside a pre-defined vocabulary, but also many abstract concepts such as wrong reasoning relating to a visual scene as well as wrong attributes of a particular objects or person. These additional hallucinations are not possible to be measured with \throne without modifications.
    \item    The presented method of \throne only focused on \emph{``Type-I hallucinations''} which does not paint a complete picture of the hallucinating behavior of an LVLM. Indeed we present POPE-C in Fig. 7 to extend hallucination measurements in both Type-I and Type-II hallucinations. We present POPE-C as an extension of POPE since we observe that POPE severely undercounts hallucinations in Type-II form. 
    \item Due to lack of general and exhaustive ground truth object label lists for a given image, our method relies on curated datasets such as MCOCO or Object365 that have detailed annotations that are complete on an image level, which are needed for our evaluation.
    \item  Our method focuses only on the hallucination bias of LVLMs but does not include measurements of other types of bias of LVLM generations (\eg related to concepts of fairness in generation) which we leave for future work. 
\end{enumerate}

\cvprsection{Ethical Considerations}\label{smsec:ethical_considerations}
% \lipsum[1]
We present \throne which is a general evaluation pipeline for measuring hallucinations (specifically ``Type-I'' hallucinations) in Large-Vision-Language Models (LVLMs). Overall we believe that our contribution is ethically positive as it measures and shows that existing public LVLMs are not yet ready to be deployed in mission critical applications, as we observe that they still suffer from hallucinating objects to a large extent. In addition we believe our presented evaluation framework also provides for a ``north-star'' in measuring evaluation and can aid the field and practitioners alike in measuring and making progress towards reducing evaluations in LVLMs as well as electing to use one LVLM over another. We note that measuring societal bias in LVLMs is highly important pre-requisite before their deployment, however this is not investigated in the current work.

\end{document}